\pdfoutput=1

\documentclass[11pt]{article}

\usepackage{acl}

\usepackage{times}
\usepackage{latexsym}
\usepackage[T1]{fontenc}
\usepackage[utf8]{inputenc}
\usepackage{microtype}
\usepackage{inconsolata}
\usepackage{graphicx}

\usepackage{algorithm}
\usepackage{algpseudocode}
\usepackage{tikz}
\usepackage{amsmath}
\usepackage{amsfonts}
\usepackage{amssymb}
\usepackage{amsthm}
\usepackage{capt-of}
\usepackage{xspace}
\usepackage{multirow}
\usepackage{booktabs}
\usepackage{subcaption}
\usepackage{cancel}
\usepackage{outlines}

\usepackage{listings}
\usepackage{wrapfig}

\usepackage{fontawesome5}
\usepackage{todonotes}

\usepackage{bbm}
\usepackage{cleveref}
\usepackage{adjustbox}
\usepackage[inline]{enumitem}

\usepackage{comment}
\usepackage{stmaryrd,scalerel}

\definecolor{changesColor}{rgb}{1, 0, 0}
\newcommand{\newterm}[1]{\textbf{#1}}
\makeatletter
\newcommand{\changes}[1]{
  \begingroup
  \color{changesColor}
  \ifmmode
    #1
  \else
    \textcolor{changesColor}{#1}
  \fi
  \endgroup
}
\makeatother

\newcommand{\cites}[1]{\citeauthor{#1}'s (\citeyear{#1})\xspace}

\newcommand{\signature}{\ensuremath{\sigma}\xspace}
\newcommand{\signatureEstimator}{\ensuremath{\widehat{\sigma}}\xspace}

\newcommand{\convergencethreshold}{\ensuremath{\epsilon}\xspace}

\definecolor{darkgreen}{HTML}{0B753C} \newcommand{\predictor}[1]{\textit{#1}\xspace}
\newcommand{\dataset}[1]{#1\xspace}

 \newcommand{\mymacro}[1]{{\ensuremath{\color{black} #1}}}
 \newcommand{\alphabet}{\mymacro{\Sigma}}
  
 \newcommand{\model}{\mymacro{q}}
 \newcommand{\truth}{\mymacro{p}}
 
 \newcommand{\str}{\mymacro{\boldsymbol{y}}}

 \newcommand{\sig}[1]{\mymacro{\signature_{#1}}\xspace}

  \newcommand{\logfrequency}{Log freq.\xspace}
  
  \newcommand{\mlu}{MLU\xspace}
  \newcommand{\concreteness}{Conc.\xspace}
  \newcommand{\nchars}{\#chars\xspace}
 
 \newcommand{\childes}{CHILDES\xspace}
 \newcommand{\babylm}{BabyLM\xspace}
 \newcommand{\unified}{Unified\xspace}

  \newcommand{\corpuspositive}{\sig{\mathrm{+}}}
  \newcommand{\corpusnegative}{\sig{\mathrm{-}}}
  \newcommand{\corpuscombined}{\sig{\mathrm{\pm}}}
  \newcommand{\intpositive}{\sig{\mathrm{I+}}}
  \newcommand{\intnegative}{\sig{\mathrm{I-}}}
  \newcommand{\intcombined}{\sig{\mathrm{I\pm}}}
  \newcommand{\extpositive}{\sig{\mathrm{R+}}}
  \newcommand{\extnegative}{\sig{\mathrm{R-}}}
  \newcommand{\extcombined}{\sig{\mathrm{R\pm}}}

\newcommand{\sighat}[1]{\mymacro{\widehat{\signature}_{#1}}}

  \newcommand{\corpuspositiveEstimator}{\sighat{\mathrm{+}}\xspace}
  \newcommand{\corpusnegativeEstimator}{\sighat{\mathrm{-}}\xspace}
  \newcommand{\corpuscombinedEstimator}{\sighat{\mathrm{\pm}}\xspace}
  \newcommand{\intpositiveEstimator}{\sighat{\mathrm{I+}}\xspace}
  \newcommand{\intnegativeEstimator}{\sighat{\mathrm{I-}}\xspace}
  \newcommand{\intcombinedEstimator}{\sighat{\mathrm{I\pm}}\xspace}
  \newcommand{\extpositiveEstimator}{\sighat{\mathrm{R+}}\xspace}
  \newcommand{\extnegativeEstimator}{\sighat{\mathrm{R-}}\xspace}
  \newcommand{\extcombinedEstimator}{\sighat{\mathrm{R\pm}}\xspace}

 \newcommand{\word}{{\mymacro{\boldsymbol{w}}}\xspace}

 \newcommand{\context}{\ensuremath{\mymacro{\boldsymbol{c}}}\xspace}
 
\newcommand{\kleene}[1]{{\mymacro{#1^*}}}
\newcommand{\defeq}{\mathrel{\stackrel{\textnormal{\tiny def}}{=}}}

\newcommand{\modelcontext}{\mymacro{\overrightarrow{\model_{\kappa}}}}
\newcommand{\empiricalmodelcontext}{\mymacro{\widetilde{\model}_{\kappa}}}

\newcommand{\pchars}{\mymacro{p}}
\newcommand{\pk}{\mymacro{p_\kappa}}
\newcommand{\q}{\mymacro{q}}

\newcommand{\prefixLanguageOp}[1]{\overrightarrow{#1}}
\newcommand{\truthword}{{\mymacro{\prefixLanguageOp{\pchars}}}}
\newcommand{\truthcontext}{{\mymacro{\prefixLanguageOp{\pk}}}}
\newcommand{\modelword}{{\mymacro{\prefixLanguageOp{\q}}}}

\newcommand{\refmodel}{{\mymacro{r}}}
\newcommand{\refword}{{\mymacro{\overrightarrow{r}}}}

\newcommand{\contexts}{\mymacro{\mathcal{C}}}

\DeclareMathOperator*{\argmin}{argmin}

\newcommand{\aoa}{\mymacro{AoA}\xspace}

\definecolor{verylightgray}{rgb}{0.95, 0.95, 0.95}

\definecolor{DarkGreen}{RGB}{0, 100, 0}

\crefname{figure}{Fig.}{Figs.}
\crefname{table}{Table}{Tables}
\crefname{appendix}{App.}{Apps.}
\crefname{section}{\S}{\S\S}
\crefformat{section}{\S#2#1#3}
\crefname{equation}{Eq.}{Eqs.}
\crefname{algorithm}{Alg.}{Algs.}
\crefname{algocf}{Alg.}{Algs.}

\title{A Distributional Perspective on Word Learning in Neural Language Models}

\author{Filippo Ficarra${}^1$\quad\quad Ryan Cotterell${}^1$ \quad\quad Alex Warstadt${}^{1,2}$\\
 ${}^1$ ETH Z{\"u}rich \quad \quad ${}^2$ University of California San Diego \\
  $\{$\texttt{\href{mailto:fficarra@ethz.ch}{fficarra},\href{mailto:rcotterell@ethz.ch}{rcotterell}}$\}$\texttt{@ethz.ch} \quad \href{mailto:awarstadt@ucsd.edu}{\texttt{awarstadt@ucsd.edu}}
  }

\begin{document}
\maketitle

\begin{abstract}
Language models (LMs) are increasingly being studied as models of human language learners.
Due to the nascency of the field, it is not well-established whether LMs exhibit similar learning dynamics to humans, and there are few direct comparisons between learning trajectories in humans and models.
Word learning trajectories for children are relatively well-documented, and recent work has tried to extend these investigations to language models.
However, there are no widely agreed-upon metrics for word learning in language models.
We take a distributional approach to this problem, defining lexical knowledge in terms of properties of the learned distribution for a target word.
We argue that distributional signatures studied in prior work fail to capture key distributional information. 
Thus, we propose an array of signatures that improve on earlier approaches by capturing knowledge of both where the target word can and cannot occur as well as gradient preferences about the word's appropriateness.
We obtain learning trajectories for a selection of small language models we train from scratch, study the relationship between different distributional signatures, compare how well they align with human word learning trajectories and interpretable lexical features, and address basic methodological questions about estimating these distributional signatures.
Our metrics largely capture complementary information, suggesting that it is important not to rely on a single metric.
However, across all metrics, language models' learning trajectories fail to correlate with those of children.\looseness=-1

\vspace{0.4em}

\includegraphics[width=1.5em,height=1.5em]{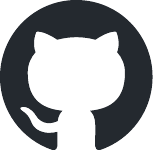}
\hspace{.6em}
\parbox{\dimexpr\linewidth-2\fboxsep-2\fboxrule}{
\vspace{-.6em}
\href{https://github.com/FilippoFicarra/word_learning}{\texttt{\nolinkurl{FilippoFicarra/word\_learning}}}
}

\end{abstract}

\section{Introduction}

There is a long tradition of characterizing words in terms of their statistical properties \citep{wittgenstein1953philosophical}.
The \newterm{distributional hypothesis} \citep{harris1954distributional,lenci2008distributional}, which characterizes knowledge of a word in terms of ``the company it keeps'' \citep{firth1957studies}, has proven surprisingly prescient.
Indeed, such a characterization is the idea behind static word representations \citep{lsi1990,landauer1997solution,hofmann1999probabilistic,mikolov2013distributed,pennington2014glove} estimated from data, as well as modern (large) language models \citep{openai2022chatgpt,ai@meta2024llama}.
While such a distributional approach to training language models (LMs) is now well-established, only recently has distributional information been explored as a tool for \emph{evaluating} lexical knowledge in LMs.\looseness=-1

Over the last few years, there has been a growing interest in studying word learning in language models \citep{nikolaus2021evaluating,chang2022acquisition,portelance2023learning,Portelance2023predicting,vong2024grounded,zhuang2024visual,zhuang2024lexicon,ma2024babysit}.
Most of these studies are part of a larger research program to use LMs to inform the study of human language acquisition by serving as convenient, controllable, and effective models of human development \citep{dupoux2018cognitive,linzen2019can,warstadt2022what, constantinescu2024exploring}.
From this perspective, it is desirable to have LMs with human-like learning trajectories, as they can better serve as generalizable models of human learners.
Word learning has a potentially important role in the success of this research program because it is one of the best proving grounds for comparing the learning trajectories of humans and LMs head to head.
While some studies \citep[e.g.,][]{choshen2021grammar} have tracked syntax learning in LMs using benchmarks like BLiMP \citep{warstadt2020blimp}, corresponding data for children is more limited in scope \citep{evanson2023language}.
There is also child data on phonological learning \citep{lavechin2022reverse} which can be explored further as audio-based LMs improve.\looseness=-1

Fortuitously, word learning trajectories in text-based LMs can be easily compared against a wealth of child data in multiple languages thanks to the massive efforts of caregivers and scholars who report and curate child word learning data in the Wordbank database \citep{frank2017Wordbank}.
Unfortunately, the caregiver reporting approach \citep{fenson2007macarthur} used in Wordbank is not immediately applicable to LMs, and there is no consensus on how to benchmark word learning in LMs.
\citet{zhuang2024visual} explored word learning through different methods, including comparing LMs’ word similarity scores to humans’ \cite{finkelstein_placing_nodate,bruni-etal-2012-distributional, hill-etal-2015-simlex, gerz-etal-2016-simverb}, classifying lexical entailment relations \cite{santus-etal-2016-cogalex}, predicting semantic features \cite{buchanan_english_2019} and using minimal pairs to measure LM preferences for appropriate word usage \citep{marvin2018targeted}.
Other articles rely on visual stimuli to ground evaluations for multimodal models \citep{nikolaus2021evaluating,berger2022computational,vong2024grounded}.
Notably, \citet{chang2022acquisition} and \citet{Portelance2023predicting} take a distributional approach, characterizing lexical knowledge in terms of the LM's surprisal, an information-theoretic quantity that has been widely studied in psycholinguistics \citep{hale2001probabilistic,levy2008expectationbased}.\looseness=-1

In this study, we take inspiration from \cites{chang2022acquisition} approach to tracking the model's distributional knowledge about a particular word throughout training.
We formalize their approach and improve on it in several respects.
While \citeauthor{chang2022acquisition} only consider the surprisal under an LM of a word in a context where the target word is appropriate and (implicitly) rely on a trivial approximation of the ground truth distribution in evaluating the quality of lexical knowledge, in contrast, we propose a family of distributional signatures that allow for the consideration of the LM's learned distribution in both appropriate and inappropriate contexts.
We also introduce distributional signatures that are truly intrinsic to the model itself, as well as strongly reference signatures that compare the learned distribution to a non-trivial ground truth, which we approximate using a large pretrained LM.\looseness=-1

In our experiments, we train language models from scratch on three datasets resembling the input to children to varying degrees.
We record the distributional signatures for a set of common words throughout training, and following \citeauthor{chang2022acquisition}, we apply a threshold to the measured learning trajectories to obtain an age-of-acquisition (\aoa) for each word.
We then conduct analyses to answer the following questions:
\begin{enumerate}
    \item Which methods allow us to reliably extract \aoa scores?
    \item How does the order of word acquisition in LMs compare to that of children?
    \item What are the empirical properties of the learning trajectories for different distributional signatures?
\end{enumerate}

We find that the learning trajectories for different distributional signatures are indeed different from each other, suggesting that earlier approaches failed to capture some aspects of word learning.
While many signatures, like \cites{chang2022acquisition}, give trajectories that are highly correlated with simple features like lexical frequency, other signatures 
are harder to predict and therefore may capture more nontrivial information.
However, we find that learning trajectories for some distributional signatures fail to converge, making \aoa{s} difficult to infer.
Finally, no signature yields \aoa scores that are strongly correlated with children's \aoa, supporting the conclusion that with current methods, LMs' learning patterns are poorly aligned with humans' and underscoring a limitation of current LMs as models of human development.
We, therefore, call for future work to evaluate and improve the human-likeness of LMs' learning trajectories using the distributional signatures we propose.

\looseness=-1

\section{Preliminaries}\label{sec:preliminaries}
Let $\alphabet$ be an \newterm{alphabet}, a finite, non-empty set of characters, e.g., Unicode symbols.\footnote{Note that most modern language models operate over tokens rather than characters.
For simplicity, our presentation is in terms of characters.} 
A \newterm{string} is a finite sequence of characters drawn from an alphabet $\alphabet$.
The set $\kleene{\alphabet}$, the Kleene closure of $\alphabet$, is the set of all strings with characters drawn from $\alphabet$---including the empty string $\varepsilon$.
We consider two distinguished types of strings.
First, we define a \newterm{word}\footnote{Defining a word is a complex matter, and we concede to not having done it justice in this article.
As a simple example, consider the English verb \textit{to run}. If a child says, \textit{I ran}, we likely interpret this as the verb \textit{to run} and this should, ideally, be taken into account in our framework. 
Yet, under our current setup, we are not able to account for inflectional morphology. Moreover, even beyond morphological inflection, it is hard to define the notion of a word \textit{in se}; see \citet{marantz2001words} for a longer discussion. 
We leave a more accurate definition of a word to future work.
} as a character string $\word \in \kleene{\alphabet}$, which is believed to operate as a lexical item.
Second, we refer to an arbitrary character string that precedes a word as a \newterm{context}. 
We denote a context as $\context \in \kleene{\alphabet}$.\looseness=-1

A \newterm{language model} $\truth$ is a probability distribution over $\kleene{\alphabet}$.
A language model's \newterm{prefix probability} is defined as the following sum\looseness=-1
\begin{equation}
    \truthword(\str) \defeq \sum_{\str' \in \kleene{\alphabet}} \truth(\str \str').
\end{equation}
Throughout the paper, we will primarily be interested in a specific ratio of $\truth$'s prefix probabilities, which we use to define the probability of a word in a context\footnote{In our formalism, $\truthword(\word \mid \context)$ is not a probability distribution over words, i.e., a probability distribution over $\kleene{\alphabet}$.
Rather, $\truthword(\word \mid \context)$ simply represents the probability of the character string $\word$ following $\context$.
} as follows
\begin{equation}\label{eq:word-distribution}
    \truthword(\word \mid \context) \defeq  \frac{\truthword(\context \word)}{  \truthword(\context)}.
\end{equation}

We are also interested in the surprisal of a word in a context, denoted $- \log \truthword(\word \mid \context)$.\looseness=-1

Now, we derive a language model $\truth$'s \newterm{context distribution} using Bayes' rule as follows
\begin{equation}\label{eq:context-distribution}
  \truthcontext(\context \mid \word) = \frac{\truthword(\word \mid \context) \truthword(\context)}{\sum_{\context \in \kleene{\alphabet}}\truthword(\word \mid \context) \truthword(\context)}.
\end{equation}
Under the assumption that $\truth$ is of finite expected length, then 
$\sum_{\context \in \kleene{\alphabet}}\truthword(\word \mid \context) \truthword(\context)$ is always finite \citep[][\S 2.1]{opedal_role_2024}.
In contrast to $\truthword(\word \mid \context)$, $\truthcontext(\context \mid \word)$ \emph{is} a distribution over $\kleene{\alphabet}$ due to the normalization present in \Cref{eq:context-distribution}.
Complementarily, we define a word $\word$'s \newterm{negative context distribution} as\looseness=-1
\begin{equation}\label{eq:negative-context-distribution}
\!\!  \truthcontext(\context \mid \neg \word) = \frac{(1-\truthword(\word \mid \context)) \truthword(\context)}{\sum_{\context \in \kleene{\alphabet}}(1-\truthword(\word \mid \context))\truthword(\context)}.
\end{equation}
 The probability $1-\truthword(\word \mid \context)$ can be thought of as follows.
 Given that $\truthword(\word \mid \context)$ is the probability of the event that a string sampled from $\truthword(\cdot  \mid \context)$ has $\word$ as a prefix, $1-\truthword(\word \mid \context)$ is the complement of that event, i.e., it is the probability that a string sampled from $\truthword(\cdot  \mid \context)$ does \emph{not} have $\word$ as a prefix.   

\begin{table*}[]
    \centering
    \begin{adjustbox}{width=2\columnwidth}
    \begin{tabular}{lllll}
    \midrule
       & \textbf{Positive} & \textbf{Negative} & \textbf{All} \\
    \midrule
\textbf{True}& $ \displaystyle -\sum_{\context \in \kleene{\alphabet}} \truthcontext(\context \mid \word)  \log \modelword(\word \mid \context)$  & $\displaystyle -\sum_{\context \in \kleene{\alphabet}} {\truthcontext(\context \mid \neg \word)}  \log \modelword(\word \mid \context)$  & $\displaystyle -\sum_{\context \in \kleene{\alphabet}} \truthcontext(\context) \log \modelword(\word \mid \context)$  \\ 

\midrule
\textbf{Intrinsic}  & $\displaystyle -\sum_{\context \in \kleene{\alphabet}} \modelcontext(\context \mid \word)  \log \modelword(\word \mid \context)$  & $\displaystyle   - \sum_{\context \in \kleene{\alphabet}} \modelcontext(\context \mid \neg \word) \log \modelword(\word \mid \context)$  & $\displaystyle -\sum_{\context \in \kleene{\alphabet}} \modelcontext(\context) \log \modelword(\word \mid \context)$  \\
\midrule 
\textbf{Reference}& $ \displaystyle \sum_{\context \in \kleene{\alphabet}} \truthcontext(\context \mid \word)  \left|\log \frac{\modelword(\word \mid \context)}{\refword(\word \mid \context)} \right|$  & $\displaystyle \sum_{\context \in \kleene{\alphabet}} {\truthcontext(\context \mid \neg \word)} \left|\log \frac{\modelword(\word \mid \context)}{\refword(\word \mid \context)} \right|$  & $\displaystyle \sum_{\context \in \kleene{\alphabet}} \truthcontext(\context) \left|\log \frac{\modelword(\word \mid \context)}{\refword(\word \mid \context)} \right|$  \\

    \bottomrule
    \end{tabular}
    \end{adjustbox}
    \caption{Overview of all distributional signatures measured in our experiments.}
    \label{tab:signatures}
   \vspace{-10pt}
\end{table*}

In the remainder of this paper, we will distinguish three LMs: $\truth$, the underlying distribution assumed to have generated the observed strings; $\model$, a parameterized model whose parameters we estimate; and $\refmodel$, a pre-trained reference LM, potentially larger and trained on more data.
A standard method of constructing a language model $\truth$ that approximates $\model$ is maximum-likelihood estimation. 
Suppose we observe a bag of $N$ samples $\Lbag\str^{(n)}\Rbag_{n=1}^N$ where $\str^{(n)} \sim \truth$, then we choose a model $\model$ that minimizes the following cross-entropy: $- \sum_{n=1}^N \log \model(\str^{(n)})$.

\section{Defining Lexical Knowledge}\label{sec:evaluating_lexical_knowledge}

Our goal is to evaluate word learning in LMs by following the \newterm{trajectory} of a \newterm{distributional signature} for each target word throughout LM training. 
However, both in terms of trajectory extraction and signature design, there are many design choices.
In this section, we explore and discuss the implications of a range of choice points in defining the distributional signature that is tracked during training.
In \Cref{sec:trajectories}, we discuss how to extract a trajectory from a timestamped sequence of distributional signatures.\looseness=-1

\paragraph{\citet{chang2022acquisition}.}
The most direct predecessor to this work, \citet{chang2022acquisition}, considered a single distributional signature: the surprisal, under the LM, of the target words in contexts where the word occurs in a test corpus.
This is a natural quantity to track during training, as it is equivalent to the cross-entropy loss per token, restricted only to samples from a single class.
In our notation, they consider
\begin{equation}\label{eq:chang-bergen}
    \widehat{\sigma}_{+}(\word) \defeq -\frac{1}{M}\sum_{m=1}^M \log \modelword(\word \mid \context^{(m)}),
\end{equation}
where $\modelword$ is the LM being analyzed, and the contexts $\context^{(m)}$ are contexts taken from a corpus where they occur \emph{before} the word $\word$, which we refer to as \newterm{positive contexts} for $\word$.
We observe that---under the assumption that the positive contexts are sampled from the ground truth context distribution, i.e., $\context^{(m)} \sim \truthcontext(\cdot \mid \word)$---\Cref{eq:chang-bergen} is a Monte Carlo estimator of the quantity
\begin{equation}\label{eq:corpus-positive}
\sigma_{+}(\word) \defeq -\sum_{\context \in \kleene{\alphabet}} \truthcontext(\context \mid \word) \log \modelword(\word \mid \context).
\end{equation}
However, even in expectation, there is one salient manner in which \citeposs{chang2022acquisition} distributional signature misses potentially valuable distributional information about the target word: it fails to account for the LM's distributional knowledge about $\word$ in \newterm{negative contexts}, where $\word$ is not found.
Beyond this limitation, this distributional signature represents only one element in a potentially very large design space; in the remainder of this section, we also explore additional distributional signatures. 

\paragraph{Considering Negative Contexts.}

Knowing the distribution of \word requires not only knowing when the word is appropriate in context but also when it is \emph{in}appropriate.
Thus, we can instead study the LM's distribution in contexts sampled according to a word's negative context distribution $\truthcontext(\cdot \mid  \neg \word)$, i.e., the context distribution over all those contexts that occur before a word that is \emph{not} $\word$ and does not have $\word$ as a prefix; see \Cref{sec:preliminaries}.
Thus, analogously to \cref{eq:corpus-positive}, we define the following distributional signature:
\begin{equation}
\sigma_{-}(\word) \defeq -\sum_{\context \in \kleene{\alphabet}} \truthcontext(\context \mid \neg \word) \log \modelword(\word \mid \context).
\end{equation}
Again, under the assumption that negative contexts are sampled from $\truth$, i.e., $\context^{(m)} \sim \truthcontext(\cdot \mid \neg \word)$, we can derive a Monte Carlo estimator as follows
\begin{equation}
    \widehat{\sigma}_{-}(\word) \defeq -\frac{1}{M}\sum_{m=1}^M \log \modelword(\word \mid \context^{(m)}).
\end{equation}
\looseness=-1

\paragraph{All contexts.}
Rounding out this series of related signatures, we design a distributional signature that considers an LM's predictions in all---both positive and negative---contexts
\begin{equation}\label{eq:all-contexts}
\sigma_{\pm}(\word) \defeq -\sum_{\context \in \kleene{\alphabet}} \truthcontext(\context) \log \modelword(\word \mid \context),
\end{equation}
where $\truthcontext(\cdot)$ represents the unconditional distribution over contexts.\footnote{
We can define $\truthcontext(\cdot)$ by normalizing $\truthword(\cdot)$, which again, requires $\truth$ to have finite expected length \citep{opedal_role_2024}.
}
Analogously, we derive the following Monte Carlo estimator where $\context^{(m)} \sim \truthcontext(\cdot)$
\begin{equation}
    \widehat{\sigma}_{\pm}(\word) \defeq -\frac{1}{M}\sum_{m=1}^M \log \modelword(\word \mid \context^{(m)}).
\end{equation}

\paragraph{Intrinsic Signatures.}\label{par:intrinsic-signatures}
We turn to a discussion of a different class of distributional signatures. 
Rather than taking the expectation with respect to the true context distribution, $\truthcontext(\cdot \mid \word)$, we now consider an \newterm{intrinsic signature}, where we take the expectation with respect to the model $\modelcontext(\cdot \mid \word)$. 
This yields three distributional signatures, analogous to those above, which are defined in the second row in \Cref{tab:signatures}. We term these $\intpositive$, $\intnegative$ and $\intcombined$, respectively. 
We discuss the estimation of intrinsic distributional signatures in \Cref{sec:intrinsic_estimation}.\looseness=-1

\paragraph{Comparing to a reference distribution.}
\label{par:reference_signatures}
The signatures \corpuspositive, \corpusnegative, and \corpuscombined estimate the relationship between the model and the underlying distribution $\truth$. 
However, the true language model $\truth$ may not be achievable---both due to the finite training data or the model class itself.
Thus, it is also meaningful to compare $\model$ to a \newterm{reference distribution} $\refmodel$, which is assumed to be a larger LM trained on more data.
Following this intuition, we define three \newterm{reference signatures}, listed in the third row of \Cref{tab:signatures}, denoted as
\extpositive, \extnegative, and \extcombined, respectively, estimate them similarly as \cref{eq:chang-bergen}.
These reference signatures are distance metrics between the target model and the reference distribution.\looseness=-1

\section{Analyzing Trajectories}\label{sec:trajectories}

Given our goal of studying the word acquisition \emph{process} in LMs, we aim to study the \newterm{trajectory} of a signature \signature for various words throughout the training of the target LM.
However, an entire trajectory may contain too much information for some analyses.
In this section, we consider a family of statistics that can be extracted from the trajectory and review the main choice points in doing so.

\paragraph{Determining $\boldsymbol{\aoa}$ by Thresholding}
\label{par:extract_aoa}
While many statistics are possible, we focus on \newterm{age of acquisition ($\boldsymbol{\aoa}$)}, which is a single number that should be interpreted as the point at which learning has advanced to a satisfactory degree.
For human learners, \citet{Braginsky2016FromUT} define \aoa as the age at which 50\% of children are such that their caregivers report them as understanding the word.
\citet{chang2022acquisition} apply this thresholding approach to LMs.
Given a trajectory, they define the \aoa to be the first time step at which the signature reaches a threshold defined as $\tau\%$ of the way between some initial value representing the beginning of learning and some final value representing the endpoint of learning.
\footnote{
This comes with several choice points:
\citeauthor{chang2022acquisition} explore a range of values for $\tau$ and report little change in qualitative results, while \citet{ma2024babysit} do observe important differences due to the choice of $\tau$.
A naive approach to defining initial and final values would be to use the first and last observed values.

However, \citet{chang2022acquisition} select the initial value as the surprisal under a random chance baseline.\looseness=-1
}
Unfortunately, thresholding in this way is only suitable when \signatureEstimator 
exhibits (roughly) monotonic change over time.
While this is true of some signatures we consider, we find empirically that \corpuspositiveEstimator, \intcombinedEstimator,  \intpositiveEstimator, and  \intnegativeEstimator are exceptions.
Thus, we adopt a different approach to extracting \aoa{s} based on the notion of a \newterm{Cauchy sequence}. 
Intuitively, we say that the target word is learned at the point in the trajectory where the value of the signature becomes close to its neighboring points in the trajectory.
Our approach is defined formally in \cref{sec:convergence}.
For the sake of uniformity, we apply this approach to all signatures and leave an exploration of thresholding approaches for suitable signatures to future work.\looseness=-1

\begin{figure*}
    \centering
    \includegraphics[width=\linewidth]{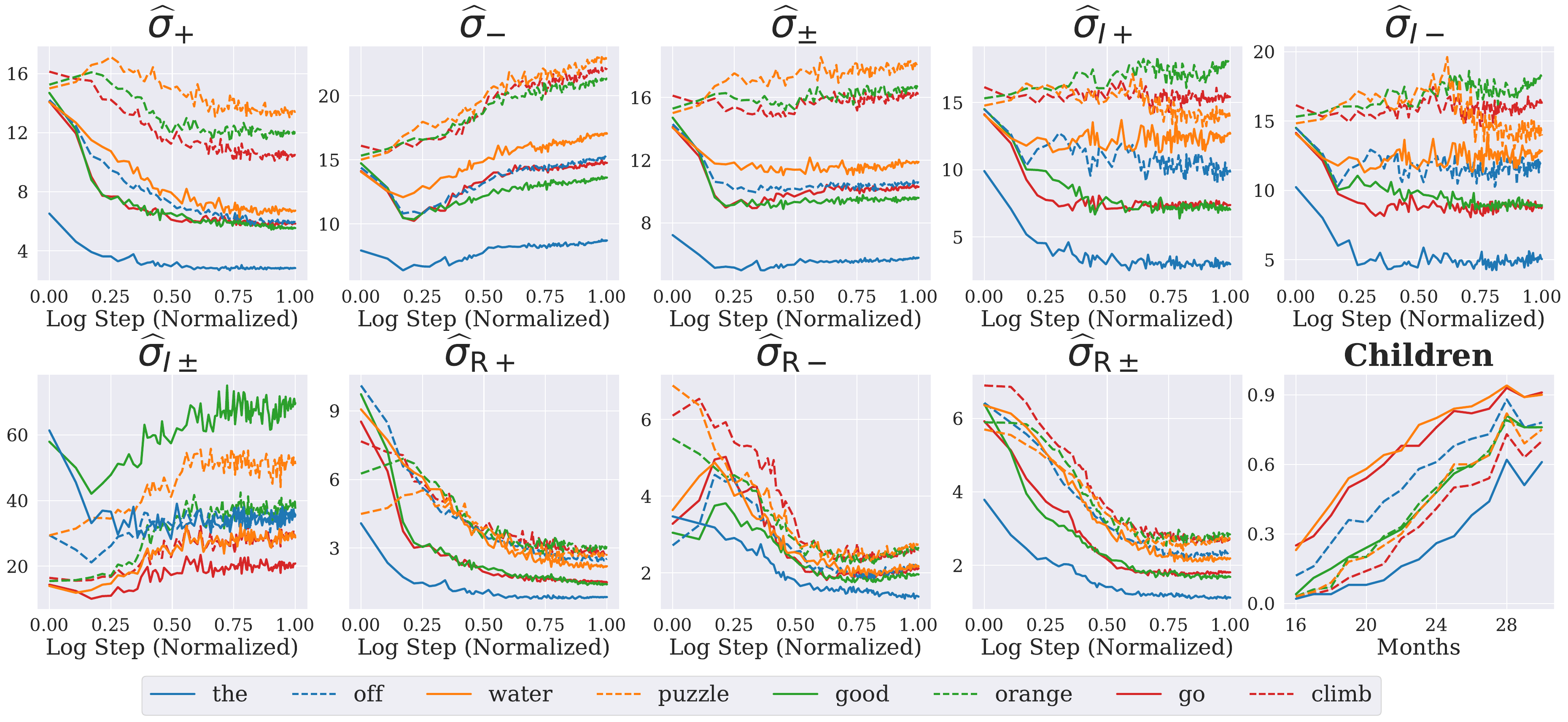}
    \caption{Trajectories for a sample of 8 words for LMs trained on the \unified dataset. We sample one high-frequency (solid line) and one low-frequency (dashed) word from each of the categories: \textsc{function words}, \textsc{nouns}, \textsc{adjectives}, \textsc{verbs}. The $y$-axis represents the value of the estimator in all $\widehat{\signature}$ plots. For the children, it represents the proportion of them that produced the word.}
    \label{fig:case_studies}
   \vspace{-15pt}
\end{figure*}

\paragraph{Smoothing the Trajectory}
Empirical trajectories may be noisy due to estimation errors or local instabilities during training \citep{datta-etal-2023-measuring}.
Thus, we consider several techniques for smoothing the trajectory.
One approach to smoothing is parametric curve fitting; \citet{braginsky_consistency} and \citet{chang2022acquisition} employ such an approach and assume trajectories follow the form of a sigmoid curve.
However, parametric curve fitting requires the modeler to assume the functional form of the curve.
If the functional form of the curve is unknown, one can instead smooth the curve using a non-parametric method, e.g., a moving average or a generalized additive model  \citep{hastie1986gam}, as done by \citet{chang2023characterizing}.
For simplicity, we opt for a moving average to smooth the trajectories in this paper.

\section{Methods}\label{sec:methods}
\subsection{Language Models}
We train several language models to explore our proposed distributional signatures.\looseness=-1
\paragraph{Training Data.}
We use three datasets previously released with train/test splits for training and evaluating our LMs.
\begin{enumerate*}[label=(\roman*)]
    \item \textbf{\unified}: This dataset was compiled by \citet{constantinescu2024exploring}. It consists of approximately 600M words sampled from a combination of three corpora: \dataset{Project Gutenberg},\footnote{\url{https://gutenberg.org}} \dataset{Wikipedia}, and \dataset{OpenSubtitles} \citep{lison-tiedemann-2016-opensubtitles2016}.
    Given that a typical 13-year-old person may be exposed to around 100M words \citep{gilkerson2017mapping}, this dataset is not fully representative of the actual input to children, although it contains a large proportion of spoken language.
    \item \textbf{\babylm}: This is the 100M text-only corpus from the second BabyLM Challenge \citep{babylm2024}. 
    The dataset is designed to be relatively developmentally plausible while also containing the amount of input that a typical adolescent is exposed to. 
    It includes child-directed speech from \dataset{\childes} \citep{macwhinney_childes_2000} and children’s stories from \dataset{Project Gutenberg} \citep{gutenberg}, as well as dialogue such as \dataset{BNC} and the \dataset{Switchboard Corpus} \citep{Stolcke_2000}, along with \dataset{Simple English Wikipedia} and \dataset{Open Subtitles}.
    \item \textbf{\childes}: This is the \childes subset taken from \babylm, consisting of 29M tokens of child-directed speech.
\end{enumerate*}
These datasets constitute an attempt to balance developmental plausibility against quantity.
Our motivation for training on datasets such as \babylm and \childes is to observe whether more developmentally plausible training distributions result in more human-like word learning trajectories.

\paragraph{Signature Estimation.}
To estimate the signatures for each word, we sample 100 positive and 100 negative contexts from the \babylm test set.
To ensure fair cross-model comparisons, we use the same test contexts for all models, regardless of training data.\looseness=-1

\paragraph{Models.}
We train GPT-2 from scratch following the training procedure described by \citet{radford2019language}.
To reduce variance in performance due to random initialization, we train three variations of each model using different random seeds.
To compute the reference signatures (\extpositiveEstimator, \extnegativeEstimator, \extcombinedEstimator) we use Llama-3.1-8B\footnote{\url{https://ai.meta.com/blog/meta-llama-3-1/}} as the reference distribution $\refmodel$.
Full details regarding the hyperparameters, training duration, and loss curves are provided in \cref{sec:training_appendix}.
As we are interested in analyzing the learning trajectories for models, it is important that they are trained for a reasonable duration.
For models trained on \babylm and \childes we apply early stopping, i.e., we choose the best model on a held-out development set, as we found that models eventually overfit.
For models trained on \unified we train for $30{,}000$ steps, or 12 epochs, following \citep{constantinescu2024exploring}.
We estimate that \citet{chang2022acquisition} trained their models on about $1.6\times10^9$ input tokens (counting repetitions).\footnote{This estimate is based on the reported 100k steps with a batch size of 128 and a context window of 128.}\looseness=-1

\subsection{The Wordbank Corpus}\label{sec:word_data}
Child \aoa data comes from the  North American English portion of the Wordbank database \cite{frank2017Wordbank}. 
For each word and month, Wordbank provides the proportion of children in the study that have produced the word by that point.
The \aoa is defined as the first month by which at least 50\% of children have produced that word \cite{goodman2008does,Braginsky2016FromUT}.
We exclude words for which we were not able to sample 100 positive context types from the \babylm dataset, leaving us with 305 words.
The words in Wordbank are divided into 4 different \textbf{lexical categories}: \textsc{Nouns} (101), \textsc{Predicates} (124), \textsc{Function Words} (45) and \textsc{Other} (49)\footnote{The total word count across all categories should be 319, but it is actually 305 due to words appearing in multiple categories.}. The  \textsc{Predicates} category is further divided into \textsc{Adjectives} and \textsc{Verbs}.\footnote{Some words, e.g., \textit{babysitter}, \textit{doctor}, and \textit{brother}, are categorized as \textsc{other} even though they belong to the category \textsc{nouns} in the category \textsc{Other}.
We exclude such misannotated words from our analyses, yielding 262 words in total.} \looseness=-1

\section{Examining LM Learning Trajectories}\label{sec:qualitative}

Before quantitatively comparing LM and child word learning trajectories in \cref{sec:predicting_aoa}, we conduct several analyses focusing solely on LM trajectories.

\subsection{Case studies}

We perform several case studies by inspecting the learning trajectories and \aoa scores for humans and each distributional signature from \cref{sec:evaluating_lexical_knowledge}.
We analyze the trajectories and \aoa scores for LMs trained on the \unified dataset for a sample of 8 words: two \textsc{function words}, two \textsc{nouns}, two \textsc{adjectives}, and two \textsc{verbs}. 
For each category, one word is chosen from the 10 most and 10 least frequent (for the \unified dataset).\looseness=-1

\cref{fig:case_studies} shows the trajectories for these words, and \cref{tab:case_studies_aoa} gives the \aoa scores.
For most signatures, we observe that the higher-frequency word from a category has an earlier \aoa than the corresponding lower-frequency word.
We also observe that most signatures yield a wide range of \aoa scores, but others---particularly $\corpusnegativeEstimator$---show very similar (and late) \aoa{s} for all words we inspect.
\cref{tab:first_last} shows the first and last learned words for each signature.
Generally, we find that high-frequency words and function words are learned first.\looseness=-1

\subsection{Convergence behavior}\label{sec:convergence_results}

As we rely on the Cauchy criterion to extract \aoa scores, we now examine how different signatures converge.
\cref{fig:case_studies} shows that the shape of the learning trajectories varies between signatures.
Within a given signature, trajectory shapes are internally consistent to varying degrees.
As expected, the reference signatures are mostly monotonically decreasing, indicating that the probability of the word of interest under the LM becomes closer to the ground truth after more iterations.
Furthermore, for the corpus-based signatures, \corpuspositiveEstimator trajectories are decreasing, whereas \corpusnegativeEstimator are increasing.\footnote{One caveat: for the first few training steps, the trajectory sometimes moves in the other direction.
\citet{chang2022acquisition} observed this phenomenon, showing that the learned distribution initially approximates a uniform distribution followed by the unigram distribution.
After this point, the trajectories are largely monotonic.}
On the other hand, the intrinsic signatures and \corpuscombinedEstimator are not consistently increasing or decreasing.\looseness=-1

\begin{figure}
    \centering
    \includegraphics[width=\columnwidth]{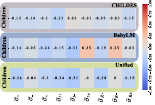}
        \caption{Pearson's correlation coefficients between different signatures and children's \aoa (C) across three datasets: \childes, \babylm, and \unified.}
    \label{fig:children_correlation}
    \vspace{-15pt}
\end{figure}

We compute the \aoa scores for a given signature using a range of values for \convergencethreshold.
\Cref{fig:number_converged} shows how many words failed to converge under different thresholds \convergencethreshold.
We find that the vast majority of trajectories converge with $\convergencethreshold=0.15$.
For lower values of $\convergencethreshold$, we see as many as half of all word trajectories failing to converge on the \childes dataset.
However, on the larger datasets \babylm and \unified, we observe high rates of convergence across the board.
Finally, it is the intrinsic signatures and \corpuscombinedEstimator that show the lowest rates of convergence.
As discussed above, these are precisely the same signatures that do not have an internally consistent shape.
Furthermore, the figures in \cref{app:thresholds_correlation} show correlations for each pair of thresholds.
With a few exceptions for extreme values, different thresholds still yield \aoa scores that are highly correlated.
Therefore, in all our results (including those discussed above), we apply an intermediate value of $\convergencethreshold=0.07$.\looseness=-1

\subsection{Comparing Signatures}\label{sec:comparing_signatures}
Another important question is whether different signatures give similar \aoa scores to each other.
\cref{app:signature_correlation} shows the correlation matrix of \aoa
scores for each signature.
First, the correlations are all notably higher for LMs trained on the \unified dataset.
Together with the finding that convergence rates are higher for this dataset, this supports the conclusion that \aoa scores become more consistent as training time increases.
We find that most pairs of signatures are weakly or negatively correlated with few exceptions.
In general, the various positive signatures (\corpuspositiveEstimator, \intpositiveEstimator, \extpositiveEstimator) have relatively strong correlations.
Across all datasets, the most strongly correlated pair is \corpuspositiveEstimator and  \extpositiveEstimator.
The negative signatures, but \intnegativeEstimator, have weak correlations with other signatures, except for the pair \corpusnegativeEstimator, \corpuscombinedEstimator which have very similar estimators.

\begin{table*}
    \centering    
    \renewcommand{\arraystretch}{1.1}
    \begin{tabular}{ccllllllc}
        \toprule
            \multicolumn{2}{c}{\bf Metadata} & \multicolumn{5}{c}{\bf Single-Predictor} & \multicolumn{2}{c}{\bf Multi-Predictor}\\
            $\boldsymbol{\aoa}$ \textbf{type} & \textbf{\#words} & \textbf{\logfrequency} & \textbf{\concreteness} & \textbf{\nchars}& \textbf{\mlu} & \textbf{Lex. cat.} & \textbf{Full} & \textbf{Full $\setminus$ \logfrequency} \\
        \cmidrule(lr){1-2}\cmidrule(lr){3-7}\cmidrule(lr){8-9}
            \it Children & \it 262 & \it 0.004 & \it 0.26 & \it -0.003 & \it 0.032 & \it 0.20 & \it \textbf{0.417} & \it 0.363 \\
            $\corpuspositiveEstimator$ &      262 & \textbf{0.614} & \textbf{0.298} &          0.135 &          0.072 & \textbf{0.304} & \textbf{0.616} & \textbf{0.392} \\
            $\corpusnegativeEstimator$ &      251 &          0.063 &          0.012 &          0.023 &            0.0 &          0.047 &          0.083 &          0.065 \\
            $\corpuscombinedEstimator$ &      245 &          0.542 &          0.265 & \textbf{0.142} &           0.04 &          0.294 &          0.546 &          0.379 \\
            $\intpositiveEstimator$ &      215 & 0.4 & 0.274 &          0.107 & \textbf{0.162} &          0.201 & 0.463 & 0.382 \\
            $\intnegativeEstimator$ &      197 &  0.234 &  0.168 &  0.028 &  0.035 &  0.126 &  0.256 &    0.179 \\
            $\intcombinedEstimator$ &      201 &        0.052 &          0.012 &          0.005 &          0.006 &           0.05 &          0.063 &           0.05 \\
            $\extpositiveEstimator$ &      262 &          0.572 &          0.295 &          0.118 & 0.088 &          0.296 &          0.582 &          0.377 \\
            $\extnegativeEstimator$  &    262 &          0.013 &            0.0 &          0.001 &          0.007 &          0.003 &          0.033 &          0.013 \\
            $\extcombinedEstimator$ &      256 &          0.292 &          0.159 &          0.083 &           0.04 &          0.122 &            0.3 &            0.2 \\
        \bottomrule
    \end{tabular}
\caption{Summary of model results for Child \aoa and LMs trained on the \unified dataset. Note: \predictor{lexical category} does not contain the category \textsc{other} which includes words that could be assigned to \textsc{nouns}, \textsc{predicates}, or \textsc{function words}.} 
\label{tab:regression_results_abridged}
\vspace{-15pt}
\end{table*}

\section{Human vs. LM Learning Trajectories}\label{sec:predicting_aoa}

We now examine the similarities and differences between word learning in LMs and humans.

\subsection{Comparing Human and LM $\boldsymbol{\aoa{s}}$}
We begin simply by measuring the Pearson correlation between human \aoa{s} and the LM \aoa{s} from each signature.
These values are plotted in \cref{fig:children_correlation}.
Overall, we observe very weak or negative correlations.
We find that the signature that correlates most changes depending on the datasets, but no correlation exceeds 0.31 (either positive or negative). 
The strongest positive correlations occur in \babylm for \intcombinedEstimator and \extnegativeEstimator, while the strongest negative correlations are from \unified and \babylm for the \intnegativeEstimator signature.\looseness=-1

\subsection{Predicting $\boldsymbol{\aoa{s}}$ from Features}

We now investigate which factors predict human and LM \aoa{s}, and compare whether these factors have similar effects.
\citet{Braginsky2016FromUT} identified several interpretable features that predict human \aoa{s}.
\citet{chang2022acquisition} previously fit linear models to predict LM \aoa{s} using these features.
We extend this analysis to our set of signatures.
Specifically, we take the \aoa scores of children and of each of our signatures as our dependent variables, and, additionally, consider the following predictors\footnote{The outcomes of regressions using single predictors can be misleading due to correlations among the predictors. Therefore, regressions with multiple predictors have been conducted as shown in the last two columns of \cref{tab:regression_results_abridged}.} that \citet{Braginsky2016FromUT} studied:
\begin{enumerate*}[label=(\roman*)]
    \item \textbf{log frequency} with respect to each LM's training dataset for LMs and with respect to \childes for children,
    \item \textbf{number of characters}, i.e., the number of symbols from $\alphabet$ in the word,
    \item \textbf{concreteness} judgments, collected from human subjects by \citet{brysbaert2014concreteness}, which indicates the extent to which a word is concrete, measured on a scale from 1 (very abstract) to 5 (very concrete),
    \item \textbf{mean length of the utterances (MLU)} with respect to each LM's training dataset for LMs and  with respect to \childes for children, and 
    \item \textbf{lexical category} \textsc{noun}, \textsc{predicate}, \textsc{function word}, and \textsc{other}, annotated by \citet{frank2017Wordbank,frank2021variability}.
\end{enumerate*}

\paragraph{Do similar factors influence LMs and Children $\boldsymbol{\aoa{s}}$?}

Regression analyses for LMs trained on \unified and for children are presented in \cref{tab:regression_results_abridged}.
For children, the adjusted $R^2$ with all predictors is $0.417$.
The strongest individual predictors of children's \aoa are \predictor{concreteness} and \predictor{lexical category}. 
\predictor{Log frequency} is a notably weak predictor on its own, though it does still contribute significant predictive power when added to a model including all other predictors.
These results largely reproduce those of \cite{Braginsky2016FromUT} and \citet{chang2022acquisition}, the latter of whom reported an adjusted $R^2$ of $0.43$ for predicting child \aoa from all features using a larger vocabulary of 571 words.
In predicting LMs’ \aoa, we identify two main patterns. First, the signatures \intcombinedEstimator and \corpusnegativeEstimator exhibit negligible relationships with any predictor. Second, among the other signatures, \predictor{log frequency} is consistently the most predictive factor, mirroring the findings of \citet{chang2022acquisition}. The next most predictive factors are \predictor{concreteness} and \predictor{lexical category}. See the figures in \cref{sec:predictors} or visualizations of \aoa{s} against each predictor.
For brevity, we focus on results from the \unified dataset. Across all signatures and datasets, language models (LMs) exhibit the opposite pattern from children when it comes to \predictor{log frequency} and \predictor{concreteness}, with more frequent words having a lower \aoa. While children tend to acquire concrete words earlier, language models appear to struggle more with processing concrete words and perform better with abstract ones.
Furthermore, although children's \aoa does not display a significant correlation with \predictor{number of characters}, most LM signatures reveal positive correlations. The exceptions are \extnegativeEstimator and \intcombinedEstimator (for \babylm and \childes), which show slightly negative correlations. Lastly, \predictor{MLU} follows a similar pattern in both children and language models.\looseness=-1

\paragraph{Does more developmentally plausible training data result in more human-like learning patterns?} 
From \cref{fig:children_correlation}, we observe that the models trained on \babylm tend to exhibit the most human-like learning trajectories according to some signatures; however, as stated above, LM trajectories are far from human-like across the board.
This finding is surprising given that \childes, which comes from discourses between caregivers and young children, most closely resembles the input received by the young children studied in Wordbank.
However, since all three datasets differ greatly in size, we cannot determine whether this result is due to data domain or dataset size.
By analyzing \cref{tab:lm_regression_summary}, we find that there is a positive effect of the training set size on the predictability of \aoa. 
We speculate that this may explain why \citet{chang2022acquisition}, who trained models on much larger datasets than ours, reported higher predictability for model \aoa scores (for \corpuspositiveEstimator). 
Additionally, we note that in the \childes dataset, \predictor{log frequency} is not significantly more predictive than other factors, in contrast to other datasets. 
Overall, the results do not exhibit any notable human-like patterns.

\section{Discussion and Conclusion}

Our main objective was to explore the space of distributional signatures of words more comprehensively, with an application to understanding word learning in LMs.
We showed that the distributional test adopted by \citet{chang2022acquisition} and \citet{Portelance2023predicting} can be viewed as an estimator of a more general distributional signature.
This insight also enabled us to define a broader family of signatures that follow a clear typology.
However, the question remains: which of these evaluations should be the focus of researchers interested in studying word learning in LMs?
One of our key findings in \cref{sec:comparing_signatures} is that many of these signatures are complementary.
This is true with respect to children's \aoa{s} as well as in comparison to each other.
Arguably, considering both positive and negative contexts provides a more complete picture of the LM's distributional knowledge, and comparing the LM's distribution against an LLM allows the signature to better reflect the gradient of the ground truth distribution, which is not directly observable.
Nonetheless, each signature we propose has a clear interpretation and may be useful for specific applications, though extracting usable \aoa scores is not always feasible.

In \cref{sec:predicting_aoa} we found that we could not predict children's \aoa{s} well from any of our proposed distributional signatures.
This result might seem somewhat surprising in light of \cites{Portelance2023predicting} finding that LM surprisal improves predictions of children's \aoa{s}.
However, we note that that work uses \corpuspositiveEstimator at the end of training as a predictor of \aoa, rather than the \aoa of the model as determined by a specific distributional signature.
Our results do further corroborate \cites{chang2022acquisition} conclusions on this question, and significantly expand them to a wider variety of signatures. 
Additionally, our findings contribute to a growing body of work finding specific differences between the language learning patterns of humans and LMs in other domains \citep[e.g.,][]{evanson2023language,constantinescu2024exploring}.
On the other hand, \citet{zhuang2024visual,zhuang2024lexicon} show that multimodal LMs can exhibit more human-like learning trajectories and also introduce a novel training objective that further improves human-likeness.\looseness=-1

Future work should apply our distributional tests to these and other potentially more human-like training procedures.
Besides learning in a world grounded in sensory experience, children also learn through interaction both with the physical world and with other agents \citep{clark2018conversation,nikolaus2023communicative}.
Moreover, unlike LMs, children have constraints on production, going through one-word and two-word utterance phases \citep{bloom1970language}. 
These factors no doubt influence the kinds of words children use early in development and may account for the precedence of concrete words.
There are only a few examples of training regimes for LMs inspired by interaction \citep{lazaridou2020multiagent,nikolaus2021modeling,ma2024babysit}.
Furthermore, the reliance on stochastic gradient descent and cross-entropy loss likely skew learning trajectories in LMs in ways that are not entirely human-like.
There are many opportunities to explore more human-like LM training, and we expect that word learning will be an important evaluation of human likeness as these are explored.

Having better mapped out the space of distributional signatures of lexical knowledge, our work paves the way for comparing trajectories of language models and humans. 
Our findings provide strong empirical support that there are large differences between how human and artificial language learners develop throughout learning and draw attention to the fact that there is significant work to be done to explore pre-training methods and datasets that result in more developmentally plausible language models.

\section{Limitations}
Our study has several limitations.
First, while we are interested in the possibility that LMs can be used as cognitive models and we attempt to use developmentally plausible data, our LMs are not trained in a way that is maximally similar to how humans learn. 
They lack exposure to speech, grounding, and interaction with other agents, all of which may significantly influence word learning.
Second, while our proposed true and reference signatures are weighted by a distribution \truthcontext, we only estimate this distribution using Monte Carlo estimation.
Future work should explore whether alternative estimation techniques, such as those  based on LLMs, yield qualitatively different results.
Third, the specifics of our findings may be sensitive to our training setup. Future work should examine whether different pre-training settings yield qualitatively different results, i.e., whether our findings are robust across various setups.
Finally, our study focuses on extracting \aoa{s} from learning trajectories, but \aoa is just one statistic that can be extracted from the learning trajectory.

\section*{Acknowledgments}

We thank the reviewers and ACs for ARR whose detailed feedback led to many improvements in our work. AW was supported by an ETH Postdoctoral Fellowship.

\bibliography{custom}

\clearpage
\appendix
\renewcommand{\thefigure}{A\arabic{figure}}
\renewcommand{\thetable}{A\arabic{table}}

\setcounter{equation}{0} 
\onecolumn
\label{sec:appendix}

\section{The Intrinsic Signature}\label{sec:estimation}
We now overview the intrinsic signature in more detail, which was omitted from \Cref{par:intrinsic-signatures}.

\subsection{An Intrinsic Metric}
\label{sec:intrinsic_estimation}
We develop an intrinsic metric, i.e., a metric that does not relay, in expectation, on the true language model $\truth$.
Thus, we consider the following information-theoretic quantity that resembles \Cref{eq:corpus-positive}, but where the expectation is taken with respect to the model itself:
\begin{equation}
\begin{aligned}\label{eq:positive-conditional-entropy}
\intpositive \defeq - \sum_{\context \in \kleene{\alphabet}} \modelcontext(\context \mid \word) \log \modelword(\word \mid \context).
\end{aligned}
\end{equation}
In contrast to \citeposs{chang2022acquisition} distribution signature, \Cref{eq:positive-conditional-entropy} is not grounded in an external language model.
Thus, it measures a notion of knowledge internal to the language model itself. 
We can also, by analogy to \Cref{eq:all-contexts}, define an intrinsic metric that considers just negative contexts
\begin{equation}\label{eq:negative-conditional-entropy}
    \intnegative \defeq - \sum_{\context \in \kleene{\alphabet}} \modelcontext(\context \mid \neg \word) \log \modelword(\word \mid \context),
\end{equation}
and one that considers all contexts 
\begin{equation}
\begin{aligned}\label{eq:all-conditional-entropy}
\intcombined \defeq - \sum_{\context \in \kleene{\alphabet}} \modelcontext(\context) \log \modelword(\word \mid \context).
\end{aligned}
\end{equation}

\subsection{A Practical Estimator}
We now discuss a scheme to estimate \Cref{eq:positive-conditional-entropy}.
First, we note that, by Bayes' rule, we have
\begin{equation}
    \modelcontext(\context \mid \word) = \frac{\modelword(\word \mid \context) \modelword(\context)}{\sum_{\context \in \kleene{\alphabet}}\modelword(\word \mid \context) \modelword(\context)}.
\end{equation}
Instead, we consider the following approximation.
Given a bag of contexts  $\contexts = \Lbag\context^{(m)}\Rbag_{m=1}^M$ that proceed a word $\word$, we construct the following
empirical approximation 
\begin{equation}\label{eq:approx-empirical}
\!\!\!\empiricalmodelcontext(\context \mid \word) = \frac{\mathbbm{1}\{\context \in \contexts\} \modelword(\word \mid \context)\,\modelcontext(\context)}{\sum_{m=1}^M \modelword(\word \mid \context^{(m)})\,\modelcontext(\context^{(m)})}.
\end{equation}
Plugging \Cref{eq:approx-empirical} into \Cref{eq:positive-conditional-entropy}, we arrive at
\begin{equation}\label{eq:approx-pointwise-join-entropy}
\intpositiveEstimator \defeq - \sum_{m=1}^M  \empiricalmodelcontext(\context^{(m)} \mid \word) \log  \modelword(\word \mid \context^{(m)}), 
\end{equation}
In the limiting case, i.e., when $\contexts$ includes all of $\kleene{\alphabet}$, 
we have $\intpositiveEstimator \rightarrow \intpositive$.
Note that \Cref{eq:approx-pointwise-join-entropy} is not a standard Monte Carlo estimator as the contexts $\context^{(m)}$ may not have been drawn from $\modelcontext(\cdot \mid \word)$, but it is still consistent.
An analogous estimator can be derived for \Cref{eq:negative-conditional-entropy} and \Cref{eq:all-conditional-entropy}.\looseness=-1

\section{The Signature $\boldsymbol{\extpositive}, \boldsymbol{\extnegative}$ and $\boldsymbol{\extcombined}$ are Distance Metrics}
\label{app:distance_metric}
The \textbf{reference} signatures as introduced in \Cref{par:reference_signatures} can be easily shown to be distance metrics.
Let $x_c \defeq \log \truthcontext(\context \mid \word)$ and $y_c \defeq \log \refword(\word \mid \context)$, we can rewrite the signatures as follows:
\begin{subequations}
\begin{align}
    \extpositive &= \sum_{\context \in \kleene{\alphabet}} \truthcontext(\context \mid \word)  \left|x_c - y_c\right|\\
    \extnegative &= \sum_{\context \in \kleene{\alphabet}} {\truthcontext(\context \mid \neg \word)} \left|x_c - y_c\right|\\
    \extcombined &= \sum_{\context \in \kleene{\alphabet}} \truthcontext(\context) \left|x_c - y_c\right|.
\end{align}
\end{subequations}
Because $\truthcontext(\context \mid \word),$ $\truthcontext(\context \mid \neg \word)$ and $\truthcontext(\context)$ are all greater than zero, the expressions above represent weighted Manhattan distances, which is a known distance metric.

\section{Training Details}
\label{sec:training_appendix}
The training was conducted in parallel across 8 GPUs, with gradient accumulation \citep{accumulation2017Hermans}
set to 16 and a batch size per device of 4. As a result, our model was trained with an effective batch size of 512.

\begin{table}[!htbp]
    \centering
    \begin{minipage}{.35\linewidth}
        \begin{tabular}{lc}
            \toprule
                \textbf{Hyperparameter} & \textbf{Value} \\  
            \midrule
                \# of heads & 12\\
                \# of layers & 12 \\ 
                learning rate & 7e-4\\
                learning rate scheduler & linear \\
                precision & fp16\\
            \bottomrule
        \end{tabular}
        \caption{Training Hyperparameters for GPT-2}
        \label{tab:hyperparameters}
    \end{minipage}
    \hspace{0.05\linewidth}
    \begin{minipage}{.35\linewidth}
        \begin{tabular}{lccc}
            \toprule
             \textbf{Dataset} & \textbf{42} & \textbf{123} & \textbf{28053}\\
             \midrule
             Childes & 2,800 & 2,800 & 2,600 \\
             BabyLM & 4800 & 7200 & 6000 \\
             Childes & 30,000 & 30,000 & 30,000 \\
             \bottomrule
        \end{tabular}
        \caption{Final steps for the model trained with seeds 42, 123, and 28053.}
    \end{minipage} 
\end{table}
\noindent We saved the checkpoints used for our analysis at increasing intervals throughout the training.
Specifically, we saved checkpoints
\begin{itemize}
    \item every 50 steps for integers in $(0, 1000]$;
    \item every 200 steps for integers in $(1000, 10000]$;
    \item every 500 steps for integers in $(10000, 30000]$.
\end{itemize}

\begin{figure}[!h]
    \centering
    \includegraphics[width=\linewidth]{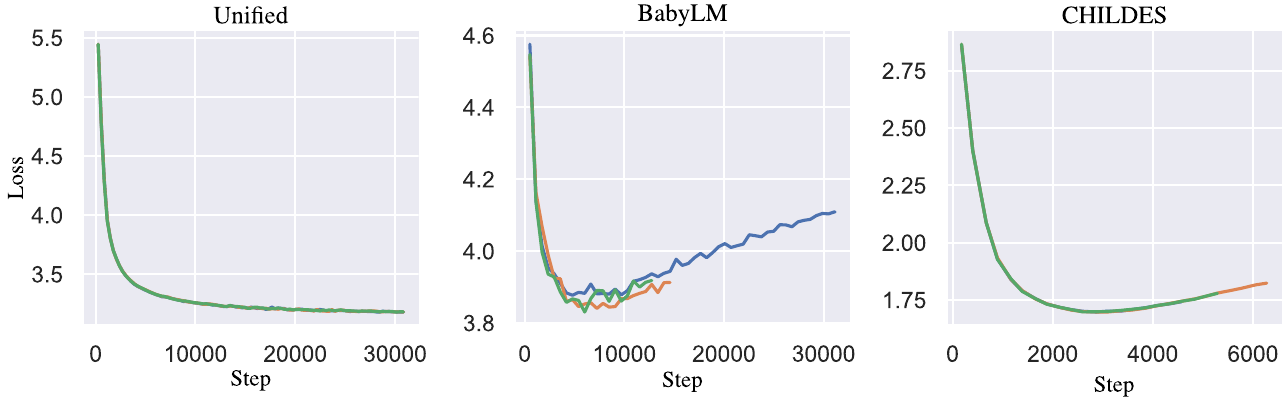}
    \caption{Validation losses for models trained on \unified, \babylm, and \childes. The curves show the necessity for an earlier stopping step for seeds 42 (blue), 123 (orange), and 28053 (green).}
    \label{fig:validation_losses}
\end{figure}

\section{Convergence with the Cauchy Criterion}
\label{sec:convergence}
To judge the convergence of learning trajectories that are non-monotonic and may not have a well-formed shape, we develop a novel technique based on the idea of a Cauchy sequence. 
Let $\signature(\word, t)$ the value that the signature of the word \word assumes at time-step $t$.
For a fixed tolerance parameter $\epsilon > 0$, the age of acquisition \aoa is defined as
\begin{equation}
\aoa = \alpha\left(\signature, \word\right
) = \argmin_{t \in \{1, \dots, T\}} \left( \max_{s, s' \in \{t, \ldots, T\}} \left|\signature(\word, s) - \signature(\word, s')\right| < \epsilon \right).\label{eq:convergence}
\end{equation}
This definition mirrors the definition of the convergence of a Cauchy sequence, albeit for a finite sequence.
However, because $T$ is finite, for small enough $\epsilon$, we do not, in general, observe true convergence in the analytic sense.
Thus, the tolerance parameter $\epsilon$ is best viewed as a hyperparameter, and our findings are dependent on the choice of $\epsilon$. 
However, given that nearly all learning algorithms are analyzed by letting $T \rightarrow \infty$, there is a sense in which our definition of AoA is well-founded.
Specifically, if we assume the convergence of the learning algorithm as  $T \rightarrow \infty$ implies the convergence of \signature, then, for every $\epsilon > 0$, there exists a number of epochs such that we will achieve $\alpha\left(\signature, \word\right)$.

\begin{figure}[!htbp]
    \centering
    \includegraphics[width=0.8\linewidth]{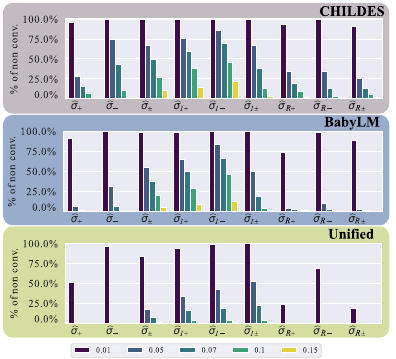}
    \caption{Percentage of words that did not converge across various \convergencethreshold.}
    \label{fig:number_converged}
\end{figure}

\Cref{fig:number_converged} shows the percentage of words that did not reach convergence for various \convergencethreshold values. A word is marked as non-converged if \signature failed to converge for even a single seed.

\section{Case Studies}
\Cref{tab:case_studies_aoa} presents the \aoa for each word illustrated in \Cref{fig:case_studies}. The \aoa values for the signatures were obtained from the model trained on the Unified dataset, using seed 42, and extracted using \convergencethreshold = 0.07.

\begin{table*}[!htbp]
    \renewcommand{\arraystretch}{1.2}
    \centering
    \begin{tabular}{lcccccccccc}
        \toprule
            \textbf{Word} & \textbf{Children} & $\boldsymbol{\corpuspositiveEstimator}$ & $\boldsymbol{\corpusnegativeEstimator}$ & $\boldsymbol{\corpuscombinedEstimator}$ & $\boldsymbol{\intpositiveEstimator}$ & $\boldsymbol{\intnegativeEstimator}$ & $\boldsymbol{\intcombinedEstimator}$ & $\boldsymbol{\extpositiveEstimator}$ & $\boldsymbol{\extnegativeEstimator}$ & $\boldsymbol{\extcombinedEstimator}$ \\ 
        \midrule
        \textit{the} & 27.79 & 0.45 & 0.94 &  0.88 & 0.49 & 0.89 & 0.89 & 0.47 & 0.80 & 0.57 \\
        \textit{off} & 22.77 & 0.68 & 0.91 & 0.82 & 0.93 & 0.95 & 0.96 & 0.67 & 0.89 & 0.67 \\
        \textit{water} & 20.00 & 0.69 & 0.91 & 0.89 & not conv. & not conv. & 0.92 & 0.68 & 0.66 & 0.67 \\
        \textit{puzzle} & 24.79 & 0.87 & 0.91 & 0.92 & 0.95 & 0.91 & 0.95 & 0.76 & 0.82 & 0.64 \\
        \textit{good} & 24.54 & 0.57 & 0.88 & 0.45 & 0.43 & 0.67 & 0.94 & 0.57 & 0.83 & 0.59 \\
        \textit{orange} & 23.26 & 0.84 & 0.92 & 0.96 & not conv. & not conv. & 0.92 & 0.80 & 0.85 & 0.64 \\
        go & 23.33 & 0.47 & 0.89 & 0.59 & 0.42 & 0.77 & 0.92 & 0.52 & 0.79 & 0.54 \\
        \textit{climb} & 26.04 & 0.73 & 0.92 & 0.93 & 0.96 & not conv. & 0.93 & 0.75 & 0.59 & 0.65 \\ 
        \bottomrule
    \end{tabular}
    \caption{\aoa for words in \Cref{fig:case_studies}.}\label{tab:case_studies_aoa}
\end{table*}

\section{First and Last Acquired Words}
\cref{tab:first_last} reports the first 10 and the last 10 words that were acquired according to each signature. The words refer to the model trained on the Unified dataset, using seed 42 and extracted using \convergencethreshold = 0.07.
\begin{table}[!htbp]
\centering
\setlength{\tabcolsep}{3pt}
\renewcommand{\arraystretch}{1.5}
\small
\begin{tabular}{lp{7cm}p{8cm}}
    \toprule
        $\boldsymbol{\ensuremath{\signature}}$ &                            \textbf{First acquired words} &  \textbf{Last acquired words} \\
    \midrule
        \corpuspositiveEstimator &          \textit{not, do, you, have, there, can, this, to, that, am} &                \textit{yes, washing, brush, toy, cow, clock, wash, puzzle, flower, egg} \\
        \corpusnegativeEstimator &              \textit{all, so, a, he, can, this, there, on, out, for} &                    \textit{red, paint, the, dinner, dry, milk, pretty, feed, cup, blue} \\
        \corpuscombinedEstimator &             \textit{under, so, like, for, all, a, at, on, out, here} & \textit{chocolate, elephant, doll, teacher, truck, gas, washing, kitchen, lips, basket} \\
        \intpositiveEstimator &     \textit{need, can, have, gonna, this, is, you, what, not, wanna} &             \textit{green, swim, touch, sleep, broken, dirty, present, park, ear, frog} \\
        \intnegativeEstimator &          \textit{have, a, take, that, so, can, here, all, need, now} &                \textit{rock, cake, money, bread, wash, cup, knife, build, teacher, sky} \\
        \intcombinedEstimator &    \textit{can, am, need, this, a, that, there, stairs, look, first} &               \textit{clock, bedroom, coat, park, your, thank, sheep, away, walk, rain} \\
        \extpositiveEstimator &       \textit{make, man, not, have, this, here, little, do, to, put} &              \textit{toy, yes, brush, flower, egg, plate, camera, star, block, washing} \\ 
        \extnegativeEstimator & \textit{truck, write, say, block, rock, hard, big, a, friend, knife} &                   \textit{arm, apple, read, present, star, buy, snow, gas, brush, slow} \\
        \extcombinedEstimator &                \textit{a, not, am, man, have, on, can, do, make, so} &                \textit{yes, apple, train, empty, frog, basket, toy, brush, draw, gonna} \\
    \bottomrule
\end{tabular}
\caption{First and last acquired words for each signature.}\label{tab:first_last}
\end{table}

\section{$\boldsymbol{\aoa}$ vs Predictors}
\label{sec:predictors}

In the following subsections, we show how log frequency, 
MLU, number of characters, and concreteness each influence \signature's \aoa across different datasets.
Each \aoa value presented in the plots represents the average \aoa across multiple seeds, extracted with \convergencethreshold = 0.07. Only words that achieved convergence across all seeds were included in the analysis.

\subsection{\childes}
{\centering
    \includegraphics[width=\linewidth]{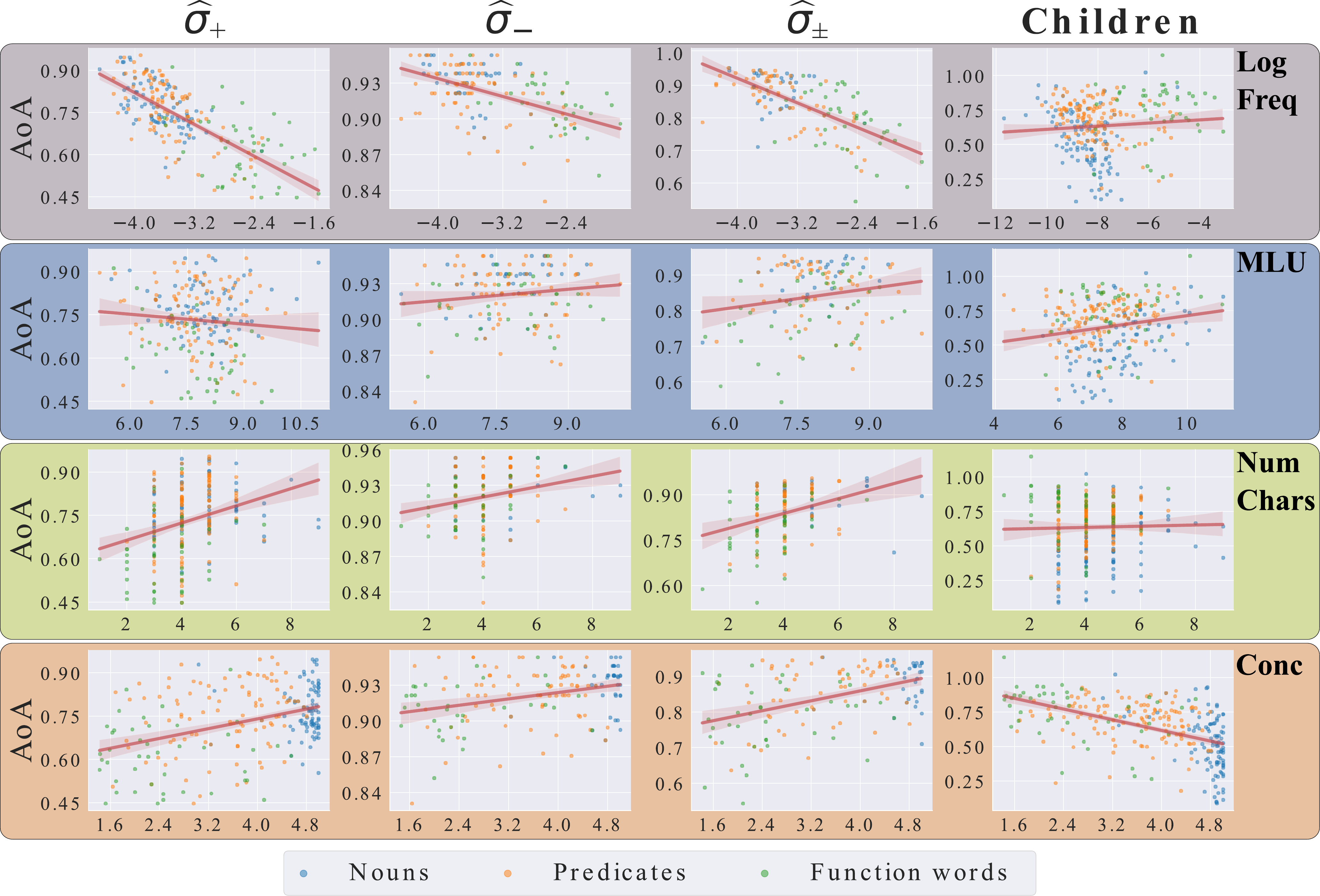}
    \vfill
    \includegraphics[width=\linewidth]{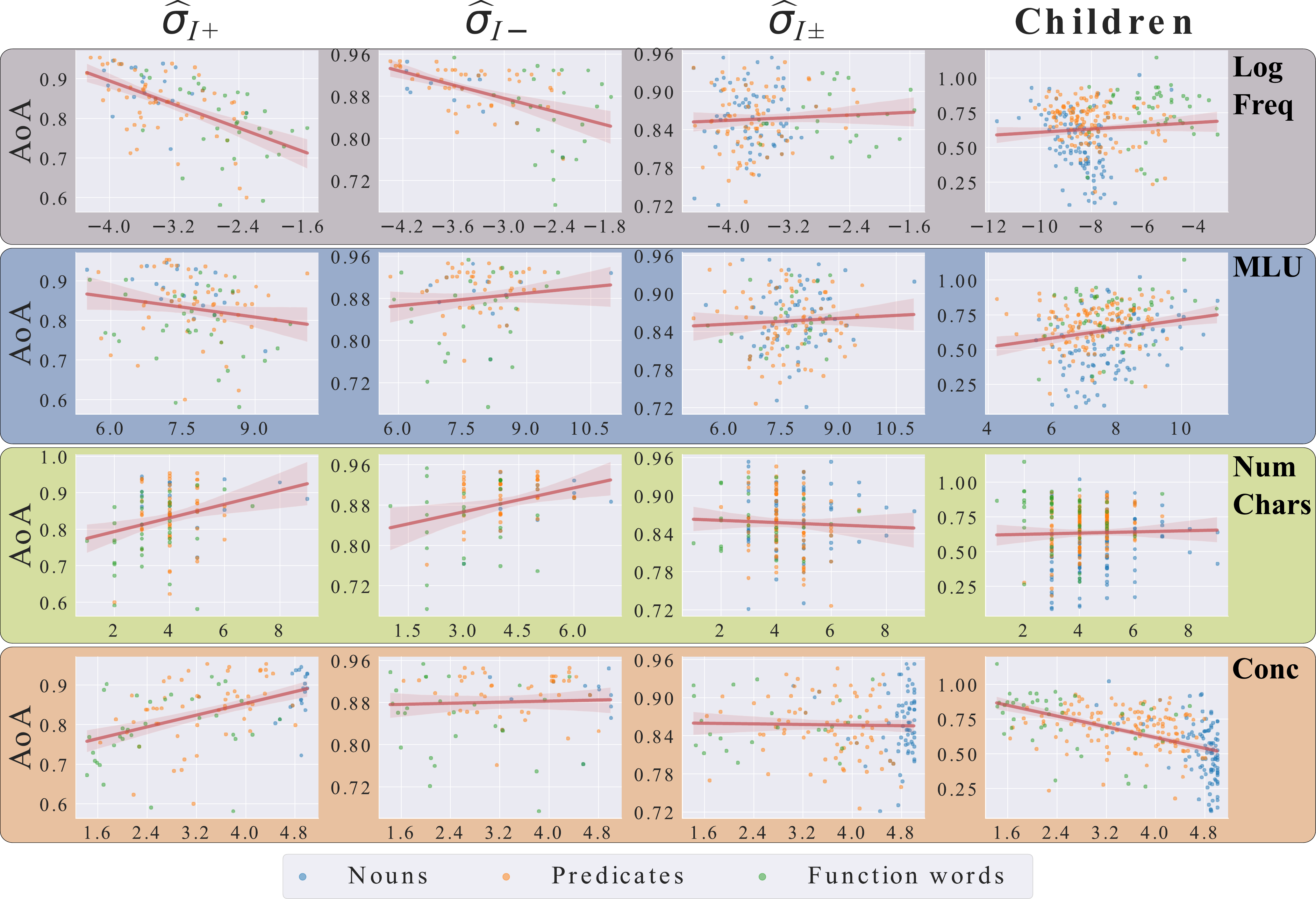}
    \vskip 1cm
    \includegraphics[width=\linewidth]{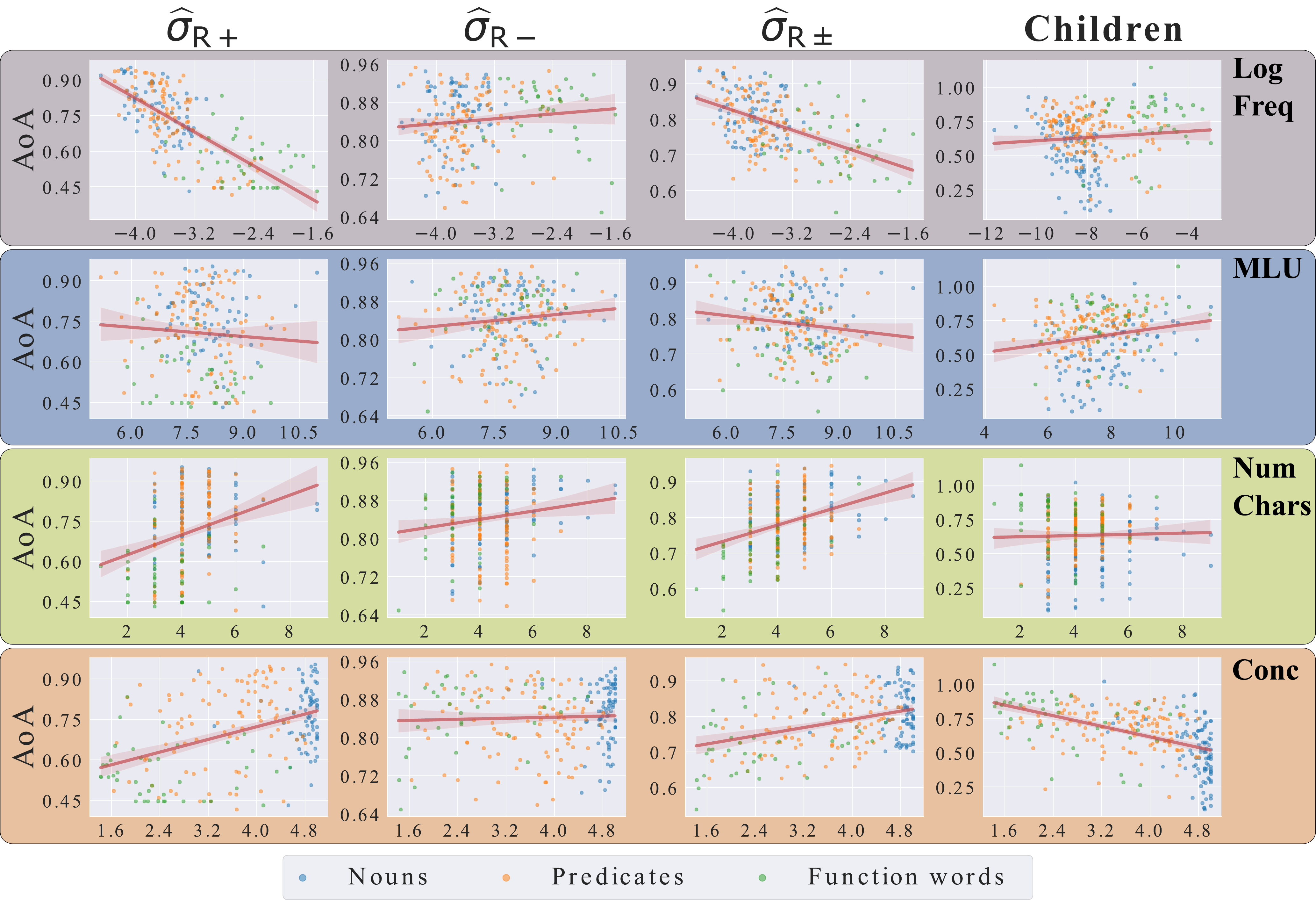}
}

\subsection{\babylm}
{\centering
    \includegraphics[width=\linewidth]{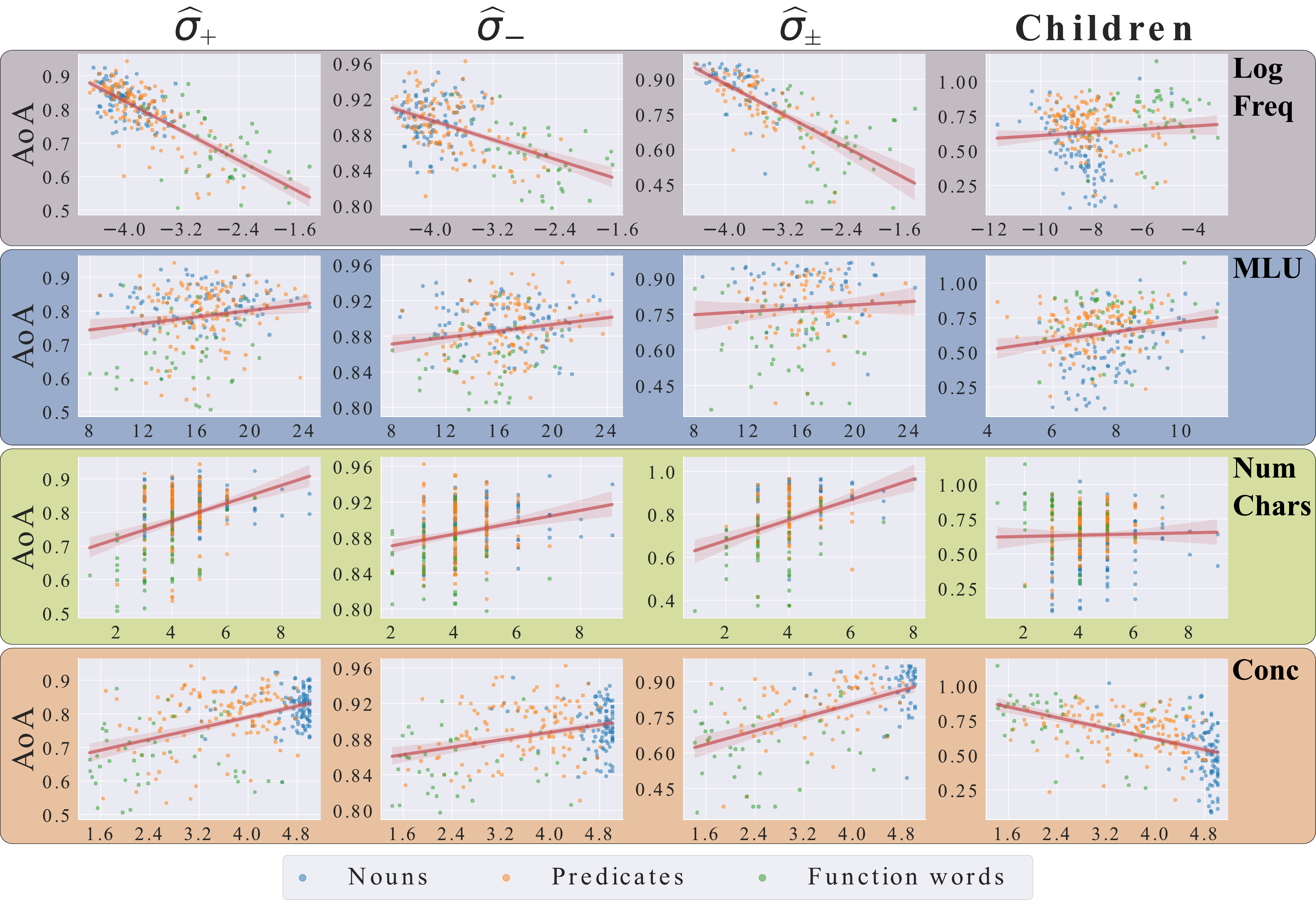}
    \vskip 1cm
    \includegraphics[width=\linewidth]{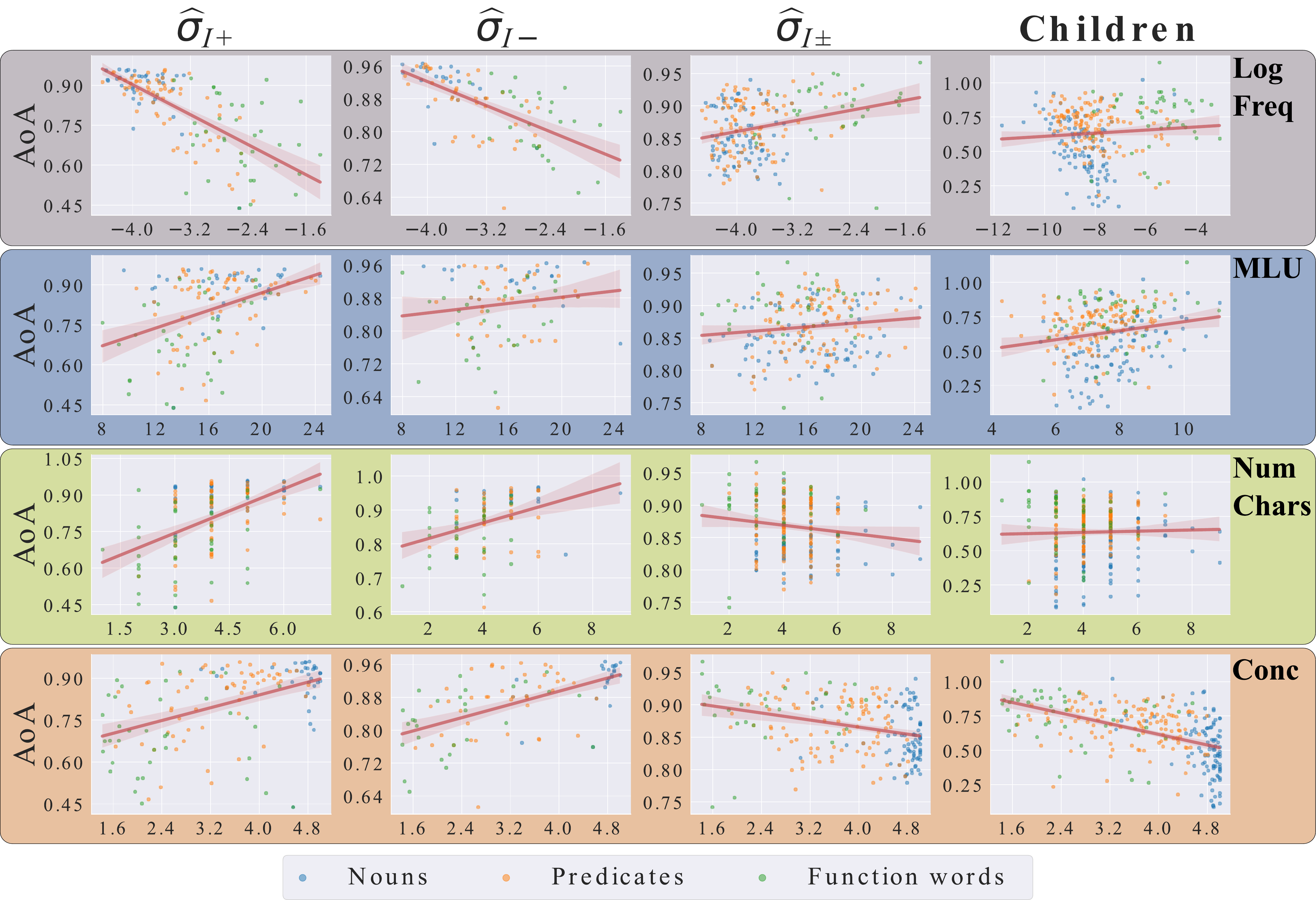}
    \vfill
    \includegraphics[width=\linewidth]{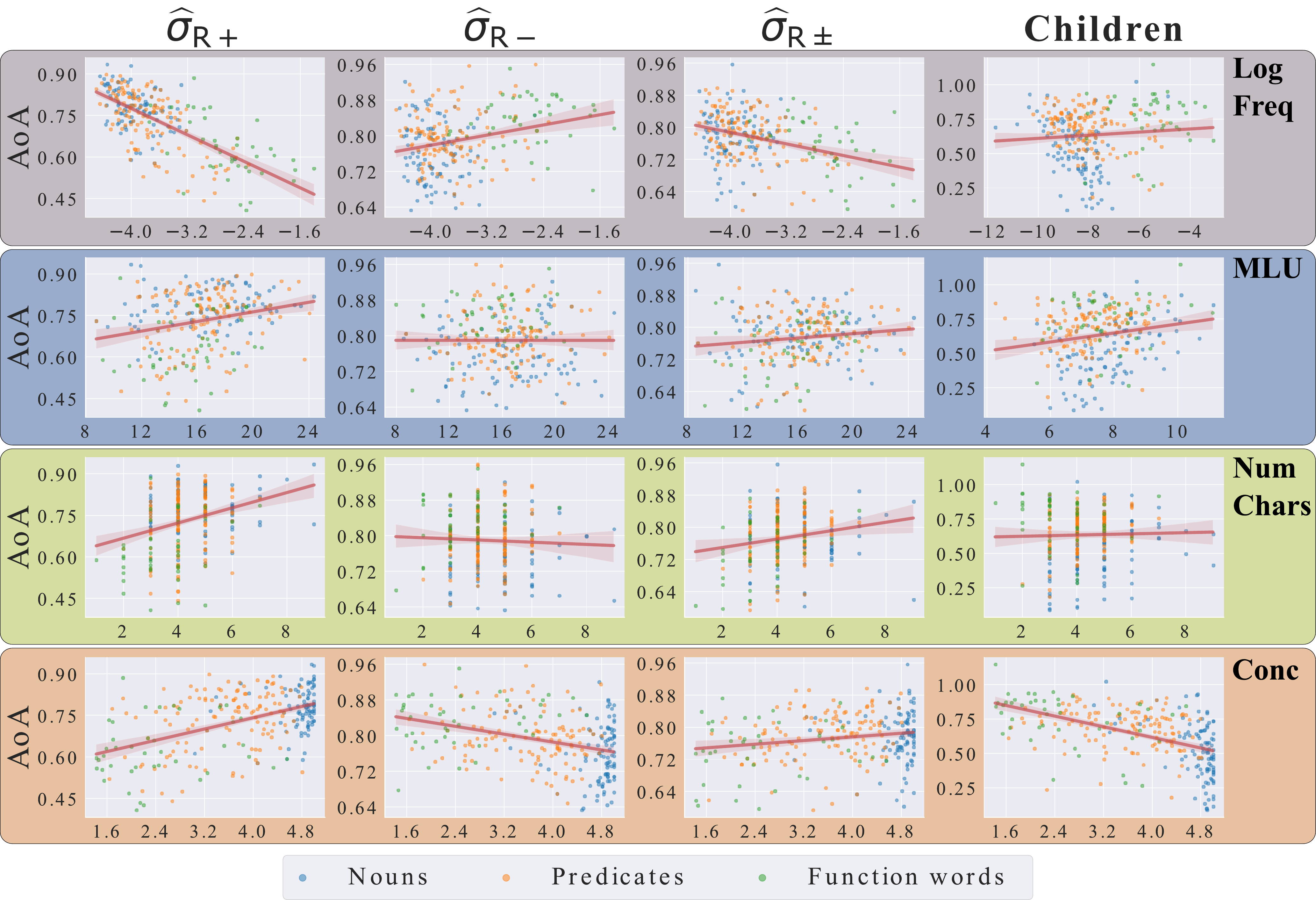}
}

\subsection{\unified}
{\centering
    \includegraphics[width=\linewidth]{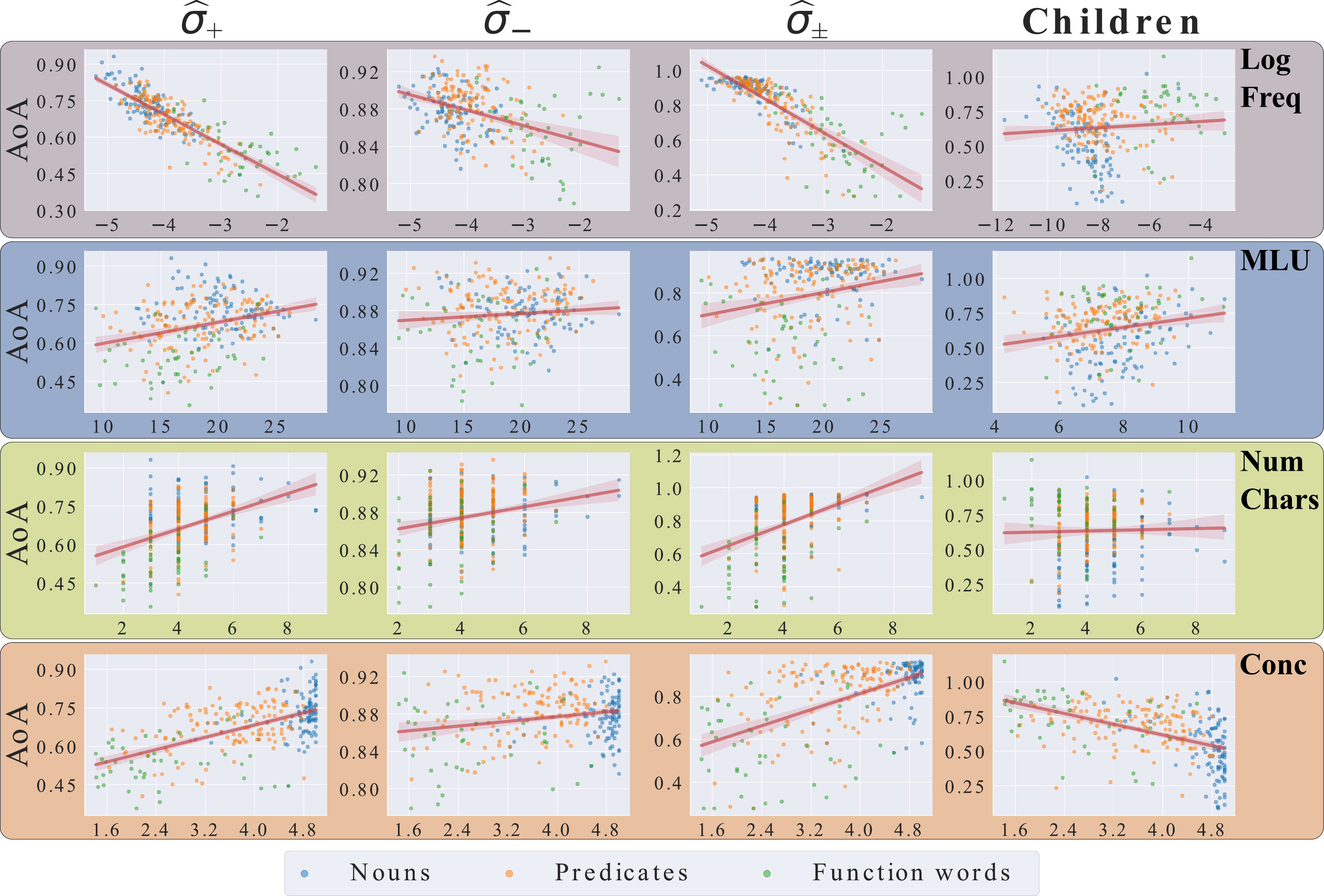}
    \vfill
    \includegraphics[width=\linewidth]{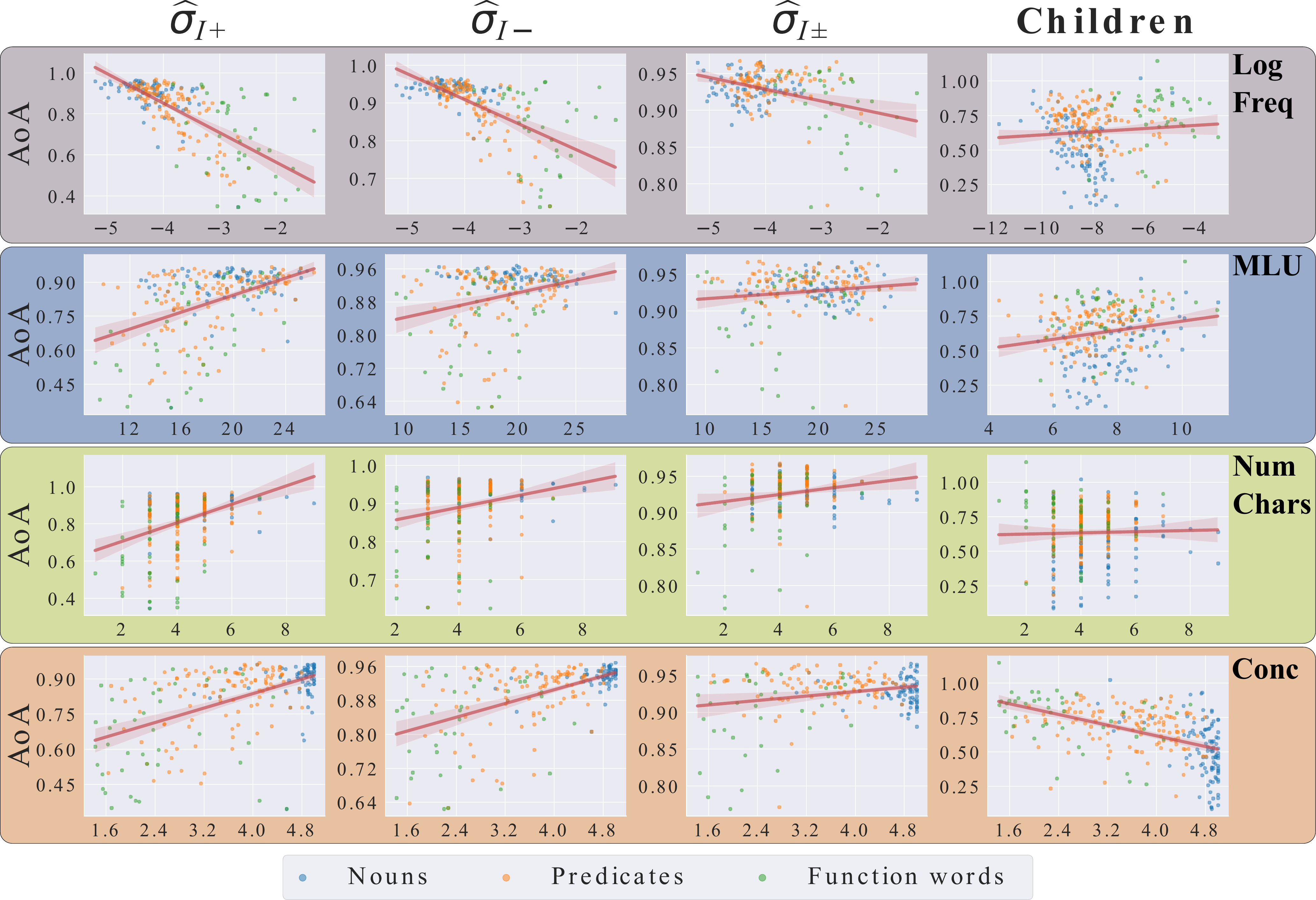}
    \vskip 1cm
    \includegraphics[width=\linewidth]{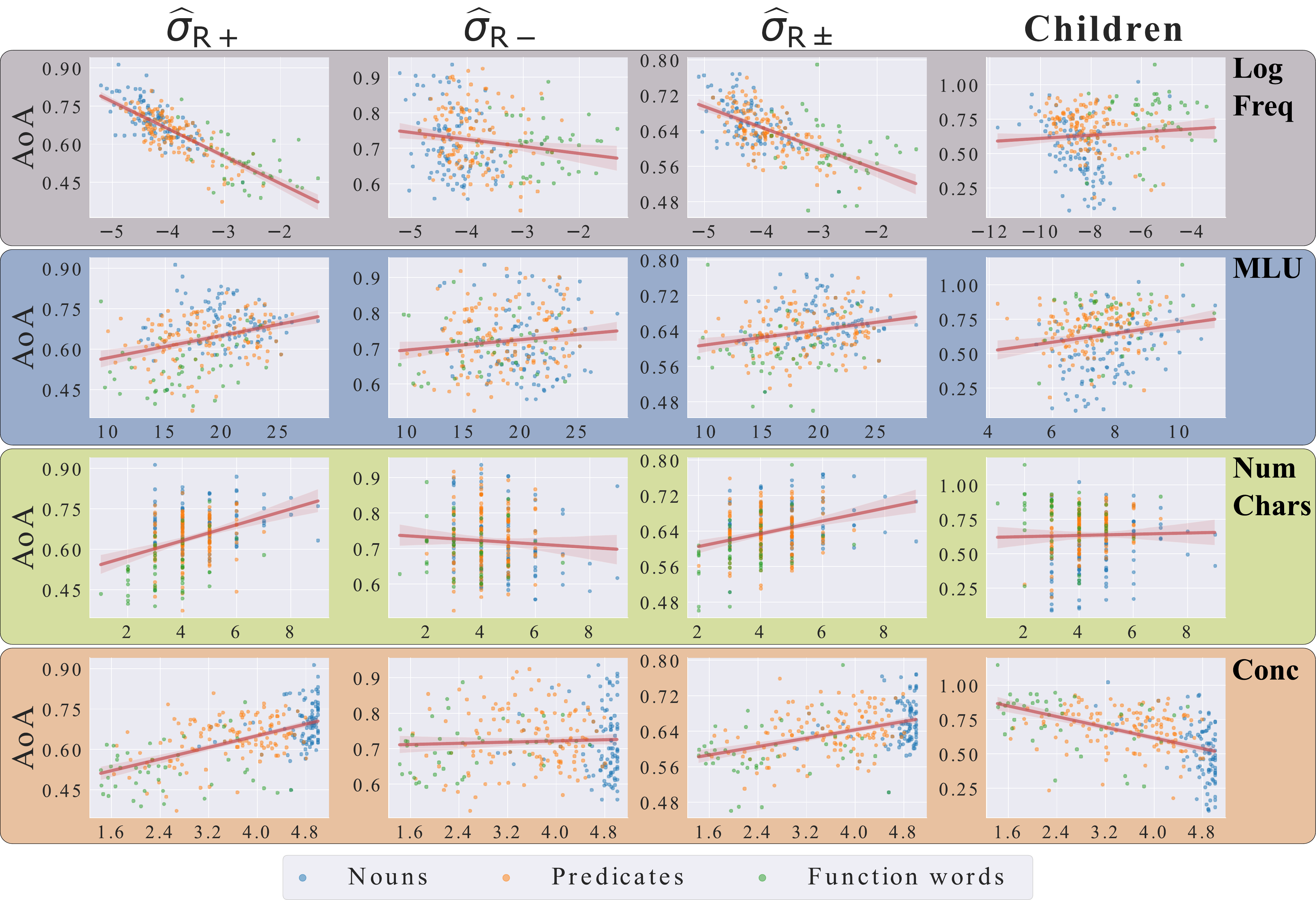}
}

\newpage
\section{Regression analysis}\label{sec:appendix_regression}

\begin{lstlisting}[caption={simplified R code for the regression analysis.}, captionpos=b]
full <- paste(predictors, collapse = "+")
reduced <- paste(original_predictors[-1], collapse = "+") 
vif_values <- vif(lm(AoA ~ full, data = data))
if (max(vif_values) > 5){
    print("Multicollinearity detected\n")
}
predictors <- c("log_frequency", "concreteness", "n_chars", "mlu", "lexical_class")
for (i in 1:length(predictors)) {
    formula <- paste("AoA ~", predictors[[i]])
    model <- lm(formula, data = data)
    cat(paste(predictors[[i]], "Adjust Rsquared:", summary(model)$adj.r.squared))
}
m_full <- lm(AoA ~ full, data = data)
m_reduced <- lm(AoA ~ reduced, data = data)
cat("Full model Adjust Rsquared:", summary(m_full)$adj.r.squared
cat("Reduced model Adjust Rsquared:", summary(m_reduced)$adj.r.squared
\end{lstlisting}
In the following regressions, we denote \predictor{log frequency} as \textbf{LF}, \predictor{concreteness} as \textbf{Co}, \predictor{number of characters} as \textbf{NC}, \predictor{mean length of utterances} as \textbf{MLU}, and \predictor{lexical category} as \textbf{LC}.
No VIF value exceeds 5, indicating that while some multicollinearity may be present, it is not severe.

\subsection{Regression on Child Learning Data}
\begin{table}[H]
    \centering
    \begin{tabular}{lllll}
        \toprule
            \textbf{Model} & \textbf{Predictor} & \textbf{Estimate} & \textbf{p-value} & \textbf{Adj. $R^2$} \\
        \midrule
            \textbf{LF} & Intercept & 0.723676 & < 2e-16 *** & 0.004 \\
           & Log Frequency & 0.011405 & 0.162 & \\
        \midrule
            \textbf{C} & Intercept & 1.003250 & < 2e-16 *** & 0.2575 \\
           & Concreteness & -0.096730 & < 2e-16 *** & \\
        \midrule
            \textbf{NC} & Intercept & 0.615790 & < 2e-16 *** & -0.003 \\
           & Number of Characters & 0.004372 & 0.661 & \\
        \midrule
            \textbf{MLU} & Intercept & 0.38715 & 1.76e-6 *** & 0.03233 \\
           & MLU & 0.03268 & 1.7e-3 ** & \\
        \midrule
            \textbf{LC} & Function Words & 0.75029 & < 2e-16 *** & 0.2012 \\
           & Nouns & -0.2284 & 6.67e-12 *** & \\
           & Predicates & -0.06586 & 0.0337 * & \\
        \midrule
            \textbf{Full} & Log Frequency & -0.047086 & 6.37e-07 *** & \textbf{0.4177} \\
           & Number of Characters & 0.025407 & 0.00342 ** & \\
           & Concreteness & -0.091746 & 3.51e-10 *** & \\
           & MLU & 0.046211 & 6.70e-08 *** & \\
           & Function Words & 0.26867 & 0.00454 ** & \\
           & Nouns & -0.193857 & 1.72e-05 *** & \\
           & Predicates & -0.084683 & 0.01456 * & \\
        \midrule
            \textbf{Full $\setminus$ LF} & Number of Characters & 0.038773 & 5.50e-07 *** & 0.3625 \\
           & Concreteness & -0.073379 & 4.91e-07 *** & \\
           & MLU & 0.043550 & 9.72e-07 *** & \\
           & Function Words & -0.042864 & 5.50e-07 *** & \\
           & Nouns & -0.115122 & 0.0085 ** & \\
           & Predicates & -0.001887 & 0.9527 & \\
        \bottomrule
    \end{tabular}
    \caption{Children regressions. For each model the Adjusted $R^2$, the estimate for the predictors, and the p-value for predictor significance.}
    \label{tab:children_regression}
\end{table}

\subsection{Regression on Language Model Learning Data}

The table below reports the Adjusted $R^2$ values for each predictor across all datasets and signatures, as introduced in \Cref{sec:predictors}. 
Consistent with previous analyses, the \aoa values were computed using a \convergencethreshold = 0.07. The table shows the number of words included in the regression analysis, counting only those that remained after outlier removal and successfully achieving convergence across all three seeds.
Among all the signatures, \corpuspositiveEstimator is the one that demonstrates the strongest predictive power across nearly all predictors.

\begin{table}[!htbp]
    \centering
    \begin{tabular}{lccccccccc}
        \toprule
            \textbf{Dataset} &  $\ensuremath{\boldsymbol{\signature}}$ &  \textbf{\#words} &    \textbf{LF} & \textbf{Co} & \textbf{NC}& \textbf{MLU} & \textbf{LC} & \textbf{Full} & \textbf{Full $\setminus$ LF} \\
        \midrule
            \newterm{\childes} &              $\corpuspositiveEstimator$ &      222 &          0.365 &          0.106 &          0.065 &          0.004 &          0.162 &          0.376 &          0.192 \\
             &              $\corpusnegativeEstimator$ &      141 &          0.108 &          0.032 &          0.025 &         -0.001 &          0.037 &          0.105 &          0.046 \\
             &            $\corpuscombinedEstimator$ &      124 &          0.236 &          0.074 &           0.06 &          0.012 &          0.081 &          0.252 &          0.129 \\
             &   $\intpositiveEstimator$ &      103 &        0.173 &          0.110 &          0.035 &          0.014 &          0.068 &          0.193 &          0.125 \\
              &    $\intnegativeEstimator$ &       77 &  0.115 & -0.003 &  0.039 &  0.003 &  0.073 &  0.132 &    0.088 \\
             & $\intcombinedEstimator$ &      163 &       -0.001 &         -0.002 &         -0.001 &           -0.0 &         -0.002 &         -0.006 &         -0.005 \\
             &    $\extpositiveEstimator$ &      210 &          \textbf{0.412} &          \textbf{0.146} &          0.069 &          0.002 &          \textbf{0.199} &          \textbf{0.422} &          \textbf{0.232} \\
              &    $\extnegativeEstimator$ &      229 &          0.006 &           -0.0 &          0.013 &          0.006 &          0.011 &          0.048 &          0.028 \\
              & $\extcombinedEstimator$ &      228 &          0.182 &          0.103 &          \textbf{0.078} &          \textbf{0.016} &          0.093 &          0.216 &          0.162 \\
        \midrule
             \newterm{\babylm}  &              $\corpuspositiveEstimator$ &      257 &           \textbf{0.48} &          0.168 &          0.092 &          0.026 &          0.217 &          \textbf{0.483} &          0.278 \\
             &              $\corpusnegativeEstimator$ &      243 &          0.124 &          0.052 &          0.029 &          0.014 &          0.083 &          0.125 &            0.1 \\
              &            $\corpuscombinedEstimator$ &      162 &          0.413 &          0.222 &          0.091 &         -0.001 &           0.29 &          0.433 &          0.323 \\
               &    $\intpositiveEstimator$ &      130 &        0.383 &          \textbf{0.224} & \textbf{0.179} &          \textbf{0.107} & \textbf{0.221} &          0.421 &          \textbf{0.363} \\
               &    $\intnegativeEstimator$ &       80 &  0.288 &  0.2 &  0.103 &  0.027 &  0.135 &  0.297 &    0.229 \\
                & $\intcombinedEstimator$ &      209 &        0.056 &          0.066 &          0.016 &          0.004 &          0.063 &          0.085 &           0.08 \\
             &    $\extpositiveEstimator$ &      253 &          0.407 &          0.204 &          0.082 &          0.057 &          0.188 &          0.425 &          0.291 \\
              &    $\extnegativeEstimator$ &      254 &          0.036 &          0.061 &           -0.0 &         -0.001 &          0.072 &          0.079 &          0.079 \\
             & $\extcombinedEstimator$ &      257 &          0.087 &          0.023 &          0.025 &          0.011 &          0.023 &          0.097 &           0.06 \\
         \midrule
             \newterm{\unified}  &              $\corpuspositiveEstimator$ &      262 & \textbf{0.614} & \textbf{0.298} &          0.135 &          0.072 & \textbf{0.304} & \textbf{0.616} & \textbf{0.392} \\
              &              $\corpusnegativeEstimator$ &      251 &          0.063 &          0.012 &          0.023 &            0.0 &          0.047 &          0.083 &          0.065 \\
             &            $\corpuscombinedEstimator$ &      245 &          0.542 &          0.265 & \textbf{0.142} &           0.04 &          0.294 &          0.546 &          0.379 \\
             &    $\intpositiveEstimator$ &      215 & 0.4 & 0.274 &          0.107 & \textbf{0.162} &          0.201 & 0.463 & 0.382 \\
              &    $\intnegativeEstimator$ &      197 &  0.234 &  0.168 &  0.028 &  0.035 &  0.126 &  0.256 &    0.179 \\
              & $\intcombinedEstimator$ &      201 &        0.052 &          0.012 &          0.005 &          0.006 &           0.05 &          0.063 &           0.05 \\
              &    $\extpositiveEstimator$ &      262 &          0.572 &          0.295 &          0.118 & 0.088 &          0.296 &          0.582 &          0.377 \\
              &   $\extnegativeEstimator$  &    262 &          0.013 &            0.0 &          0.001 &          0.007 &          0.003 &          0.033 &          0.013 \\
            & $\extcombinedEstimator$ &      256 &          0.292 &          0.159 &          0.083 &           0.04 &          0.122 &            0.3 &            0.2 \\
        \bottomrule
    \end{tabular}
    \caption{Table reporting the Adj. $R^2$ for the linear models predicting LM's \aoa.}
    \label{tab:lm_regression_summary}
\end{table}

\newpage
\section{The Impact of Convergence Thresholds on AoA Extraction}
\label{app:thresholds_correlation}

\begin{figure}[H]
    \centering
    \includegraphics[width=\textwidth]{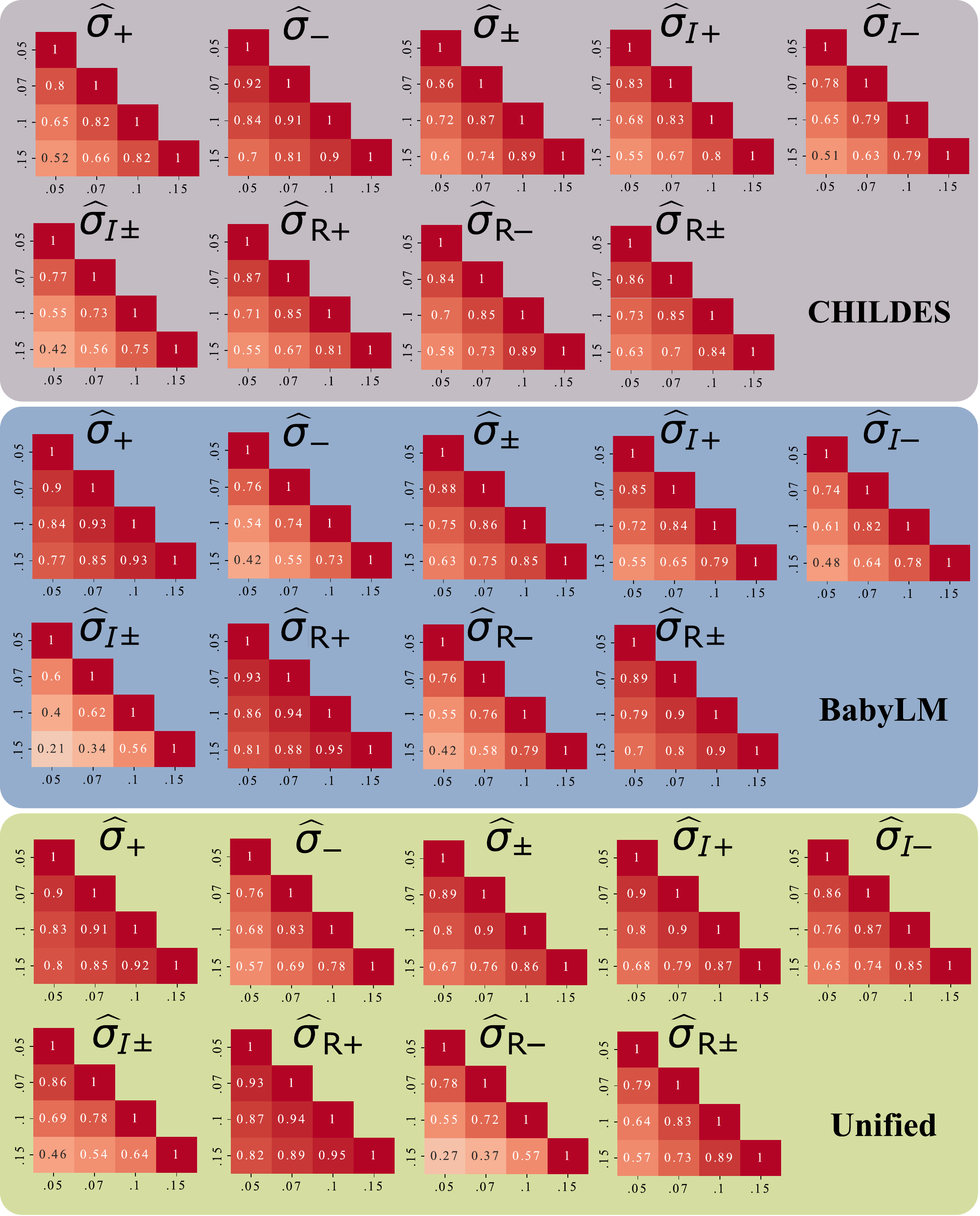}
    \caption{For each estimator, we present the Pearson correlation coefficient matrix comparing different \convergencethreshold. Warmer colors indicate stronger positive correlations, cooler colors indicate stronger negative correlations.}
    \label{fig:correlation_across_thresholds}
\end{figure}

The Pearson correlations reported in \Cref{fig:correlation_across_thresholds} illustrate how the \aoa values extracted using varying \convergencethreshold correlate. This analysis aims to determine whether the choice of \convergencethreshold significantly impacts the results. As discussed in \Cref{sec:convergence_results}, with few exceptions, the results across different \convergencethreshold values show high correlations. Therefore, our analysis will remain consistent regardless of the choice of \convergencethreshold.

\section{Correlation of Various \aoa}
\label{app:signature_correlation}
The figures in this section show how each signature's \aoa values correlate with one another across each model trained with different datasets. As a result, the dataset itself influences the correlation patterns among the different signatures. For example, the Unified dataset displays only positive correlations, whereas Childes and BabyLM datasets exhibit negative correlations for \intcombinedEstimator and \extnegativeEstimator.
The \aoa values were extracted using \convergencethreshold = 0.07.
\begin{figure}[H]
    \centering
    \begin{minipage}{0.49\textwidth}
        \centering
        \includegraphics[width=\textwidth]{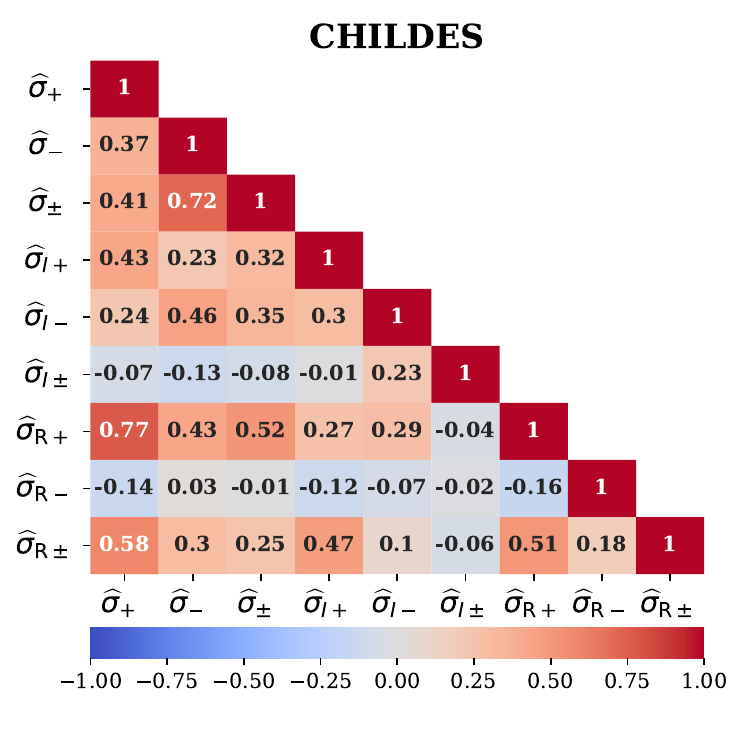}
    \end{minipage}
    \begin{minipage}{0.49\textwidth}
        \centering
        \includegraphics[width=\textwidth]{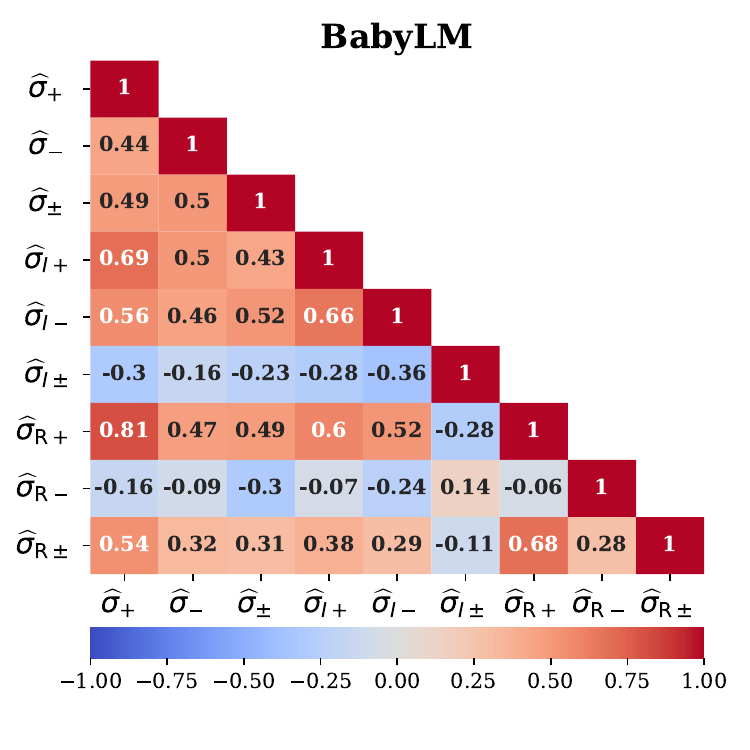}
    \end{minipage}
    \raggedright
    \begin{minipage}{0.49\textwidth}
        \centering
        \includegraphics[width=\textwidth]{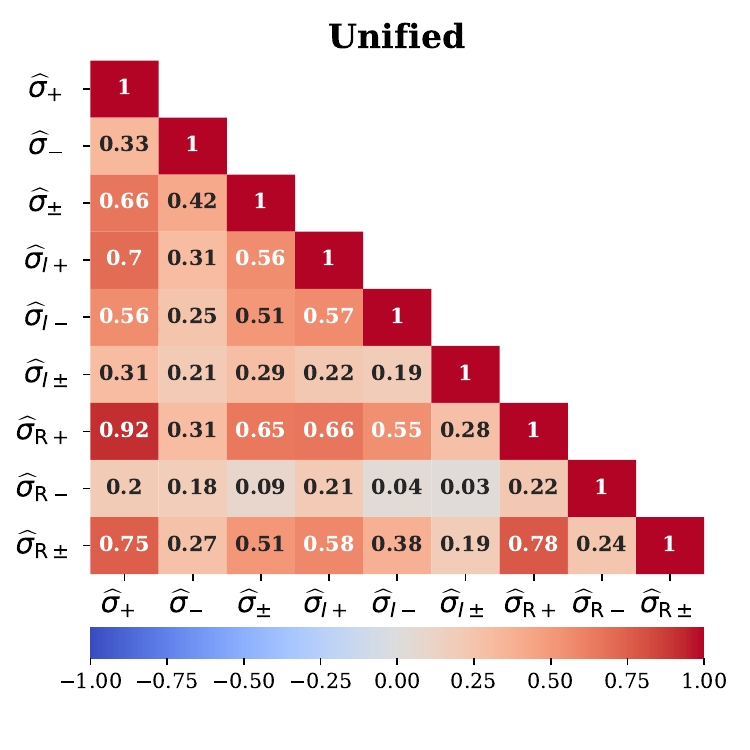}
    \end{minipage}
    \label{fig:signature_correlations}
\end{figure}

\end{document}